\documentclass[aap,preprint]{imsart}

\usepackage{xr}

\RequirePackage{amsthm,amsmath,amsfonts,amssymb}
\RequirePackage[numbers]{natbib}
\RequirePackage[colorlinks,citecolor=blue,urlcolor=blue]{hyperref}
\RequirePackage{graphicx}

\usepackage{url}
\usepackage{caption}
\usepackage{dsfont}
\usepackage{color}
\usepackage[T1]{fontenc}
\usepackage[utf8]{inputenc}
\usepackage{comment}
\usepackage{mathtools}
\usepackage{bm}
\usepackage{appendix}
\usepackage{float}

\startlocaldefs

\newtheorem{theorem}{Theorem}[section]

\theoremstyle{remark}

\endlocaldefs

\begin{document}

\begin{frontmatter}
\title{Prediction of Hereditary Cancers Using Neural Networks}
\runtitle{Prediction of Hereditary Cancers Using Neural Networks}

\begin{aug}
\author[A]{\fnms{Zoe} \snm{Guan}},
\author[B]{\fnms{Giovanni} \snm{Parmigiani}},
\author[B]{\fnms{Danielle} \snm{Braun$^*$}},
\and
\author[B]{\fnms{Lorenzo} \snm{Trippa$^*$}}
\affiliation[A]{Department of Epidemiology and Biostatistics, Memorial Sloan Kettering Cancer Center}

\affiliation[B]{Department of Biostatistics, Harvard T.H. Chan School of Public Health and Department of Data Sciences, Dana-Farber Cancer Institute}

\dedicated{$^*$These authors contributed equally.}
\end{aug}

\begin{abstract}
Family history is a major risk factor for many types of cancer. Mendelian risk prediction models translate family histories into cancer risk predictions based on knowledge of cancer susceptibility genes. These models are widely used in clinical practice to help identify high-risk individuals. Mendelian models leverage the entire family history, but they rely on many assumptions about cancer susceptibility genes that are either unrealistic or challenging to validate due to low mutation prevalence. Training more flexible models, such as neural networks, on large databases of pedigrees can potentially lead to accuracy gains. In this paper, we develop a framework to apply neural networks to family history data and investigate their ability to learn inherited susceptibility to cancer. While there is an extensive literature on neural networks and their state-of-the-art performance in many tasks, there is little work applying them to family history data. We propose adaptations of fully-connected neural networks and convolutional neural networks to pedigrees. In data simulated under Mendelian inheritance, we demonstrate that our proposed neural network models are able to achieve nearly optimal prediction performance. Moreover, when the observed family history includes misreported cancer diagnoses, neural networks are able to outperform the Mendelian BRCAPRO model embedding the correct inheritance laws. Using a large dataset of over 200,000 family histories, the Risk Service cohort, we train prediction models for future risk of breast cancer. We validate the models using data from the Cancer Genetics Network.
\end{abstract}


\end{frontmatter}

\section{Introduction}

Family history is a major risk factor for many types of cancer, including breast, colorectal, and pancreatic cancer. Various family history-based cancer risk prediction models have been developed \citep{berry1997probability, chen2006prediction, wang2007pancpro} and are used in clinical practice to guide decisions about screening and interventions. Existing models are primarily based on two approaches: 1) using Mendelian laws of inheritance to translate detailed family history information into risk predictions \citep{berry1997probability, tyrer2004breast, antoniou2004boadicea, wang2007pancpro, wang2010estimating} and 2) using summaries of family history (for example, the number of relatives with a previous cancer diagnosis) as covariates in regression models \citep{gail1989projecting, balmana2006prediction, gail2007projecting, matsuno2011projecting, banegas2017projecting, tice2008using, choudhury2020icare, choudhury2020comparative}. Recently, deep learning models based on mammographic images have also been proposed \cite{portnoi2019deep, yala2019deep}.

Mendelian models take as input a pedigree (Figure \ref{fig:pedigree}) that reflects family history of cancer (including relatives' cancer diagnoses, ages at cancer onset, and current ages). They estimate an individual's probability of carrying a mutation in a cancer susceptibility gene using Mendelian laws of inheritance, Bayes' Rule, and estimates of mutation prevalence and penetrance (probability of disease given genotype) from epidemiological literature (for example, see \cite{chen2007meta}). The individual risk of cancer is then calculated as a weighted average of mutation carrier and non-carrier risks of developing cancer. Mendelian models are typically recommended over regression-based models for individuals with a strong family history of cancer, since Mendelian models use more detailed family history information \citep{quante2012breast, pichert2003evidence}. However, they rely on explicit assumptions about cancer susceptibility genes, some of which may be unrealistic or restrictive. Known susceptibility genes account for a limited proportion of familial risk \citep{easton1999many}, and existing Mendelian models consider only a small subset of these genes. Furthermore, Mendelian models are sensitive to misreporting of family history \citep{braun2014extending, katki2006effect} and rely on accurate estimation of mutation prevalence and penetrance, which is challenging due to low mutation prevalence and heterogeneity of prevalence across populations.

The main limitations of Mendelian models can be overcome by neural networks (NNs) that eliminate the need to explicitly specify the effects of cancer susceptibility genes. A NN \citep{bishop1995neural, nielsen2015neural} is a model based on a directed graph that represents the relationship between a set of input features, typically provided in the form of a vector or matrix, and an outcome of interest. The graph consists of layers of nodes that apply a series of potentially non-linear transformations to the input to produce a prediction or classification. In our setting, the input to the NN will be a set of variables that describes the family history of an individual who presents for risk assessment. Under mild assumptions, NNs are theoretically capable of approximating any continuous function with arbitrary precision \citep{cybenko1989approximation, hornik1991approximation,leshno1993multilayer}, and in practice they have achieved state-of-the-art performance in many tasks, such as image recognition \citep{krizhevsky2012imagenet} and natural language processing \citep{hinton2012deep}. The flexibility of NNs combined with large databases can potentially lead to accuracy gains over Mendelian models. However, while the literature on NNs is extensive, little work has been done to evaluate their performance in the context of family history-based cancer risk prediction. \cite{kokuer2006comparison} trained a NN to classify families into risk categories for hereditary colorectal cancer, but they used simple summaries of family history and cross-validated their model on a relatively small dataset with 313 pedigrees. To the best of our knowledge, there is no previous work leveraging large databases of pedigrees to develop NNs for cancer risk prediction.

In this paper, we develop new NN models to predict future risk of breast cancer based on pedigree data and investigate their ability to learn patterns of inherited susceptibility. We propose a method for mapping pedigrees to fixed-size NN inputs and apply two types of NNs : 1) standard fully-connected NNs (FCNNs), and 2) convolutional NNs (CNNs) that exploit pedigree structure. Our methodological contribution is adapting CNNs for pedigree data
by defining local functions, similar to convolutional filters for image classification, that are applied repeatedly to sets of first-degree relatives within the pedigree (Section \ref{pedigreeCNNs}). We compare the performance of NNs to BRCAPRO \citep{parmigiani1998determining}, a widely used Mendelian model, and logistic regression (LR). While there are many established risk factors for breast cancer \cite{gail1989projecting, brentnall2019case}, in this paper we focus on prediction models based on family history. To allow for an interpretable comparison with BRCAPRO, which uses only family history information (along with race and ethnicity), the NN and LR models trained here do not include risk factors beyond family history (the counselee’s age and personal history of cancer are considered to be part of the family history information). The inputs to the NN models are specified in Section \ref{notation}. We note that it is straightforward to add new risk factors (e.g. breast density) to the NN models and we discuss how this can be done in the Sections \ref{mapping} and \ref{pedigreeCNNs} (the methodology for FCNNs remains identical, while adding new features to CNNs potentially requires modifying the way in which nodes are connected).  In our simulations, we generate data based on the Mendelian assumptions of BRCAPRO and determine how large a sample size is needed for NNs to achieve competitive performance compared to the generating model. Moreover, we show that when the observed family history includes misreported cancer diagnoses, NNs are able to outperform the Mendelian BRCAPRO model embedding the correct inheritance laws.

In our data application, we train NNs using over 200,000 families from the Risk Service database and validate the models on data from the Cancer Genetics Network (CGN). Although we focus on breast cancer risk prediction in our simulations and data application, the proposed approach can also be applied to other cancers. 

\section{Methods}\label{methods}

\subsection{Notation}\label{notation}

Our notation is summarized in Table \ref{table:notation} of the supplementary material (SM). Consider a counselee (someone who presents for risk assessment) who has not previously been diagnosed with a given type of cancer. Let $t$ be a pre-specified number of years. Let $Y_0=1$ if the counselee develops the cancer of interest within $t$ years and $Y_0=0$ otherwise. The goal is to estimate $P(Y_0=1|H)$, where $H$ represents family history (described below).

\begin{figure}[h!]
\centering
  \includegraphics[trim={0 1cm 0 1cm}, clip, width=0.8\columnwidth]{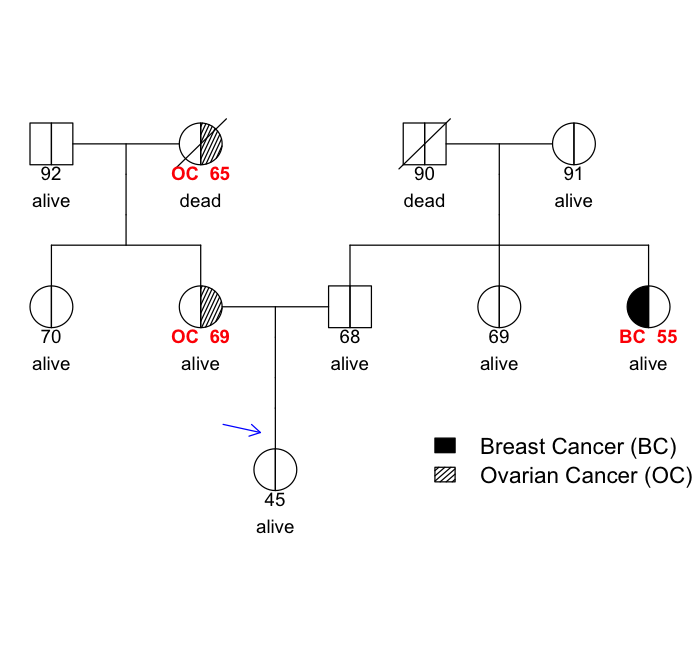}
  \caption{Example of a pedigree with family history of breast and ovarian cancers. Circles represent females and squares represent males. The arrow indicates the counselee, the individual undergoing risk assessment. Numbers below each family member represent the individual's current age if alive and unaffected, age at death if dead, and age of diagnosis if affected by breast or ovarian cancer.}
  \label{fig:pedigree}
\end{figure}

Family history can be visualized using a pedigree (Figure \ref{fig:pedigree}), a directed graph where nodes correspond to family members and edges flow from parents to offspring. The pedigree graph can be represented as a matrix $H$ where each row corresponds to a family member, containing their features and the indices of their parents. Let $R$ be the number of relatives in the pedigree besides the counselee. The family members are indexed by $r=0,1,\dots,R$, where $r=0$ corresponds to the counselee. We have $K$ features for each family member $r$: $H_{r1}, \dots, H_{rK}$. In this paper, we will consider the following $K=6$ features for breast cancer risk prediction: $H_{r1}=$ current age or age at death, $H_{r2}=$ breast cancer status (1 if affected, 0 otherwise), $H_{r3}=$ ovarian cancer status (1 if affected, 0 otherwise), $H_{r4}=$ age at onset of breast cancer (0 if unaffected), $H_{r5}=$ age at onset of ovarian cancer (0 if unaffected), and $H_{r6}=$ sex (0 if female, 1 if male). Furthermore, let $A_{r1}$ be the index of $r$'s mother and $A_{r2}$ the index of $r$'s father (either of which can be unknown). Let $H_r=(H_{r1}, \dots, H_{rK}, A_{r1}, A_{r2}) \in \mathds{R}^{K+2}$. $H$ is a matrix with $R+1$ rows and $k+2$ columns, where for $r=0,\dots,R$, row $r+1$ contains the information for family member $r$.

\subsection{Fully-Connected Neural Networks}\label{FCNNs}

A NN is a directed graph consisting of a sequence of layers (see \cite{bishop1995neural} or \cite{nielsen2015neural} for examples and  graphical representations of NNs). Each layer is a set of nodes that are linked to nodes in the previous layer through incoming edges and to nodes in the next layer through outgoing edges. A node receives a set of inputs via incoming edges, computes a function of its inputs, and propagates the result via outgoing edges. The first layer, which receives the input features (typically in the form of a vector), is called the input layer (in our setting, the input features will correspond to the family history of the counselee). The final layer, which provides the output in the form of  prediction or classification, is called the output layer. The layers in between, which are optional layers that apply transformations to the input data, are called hidden layers. A FCNN is a NN where every node in a given layer is connected to every node in the previous layer. FCNNs take as input a fixed-length vector $X$. In the context of cancer risk prediction, $X$ is a vector representation of the pedigree $H$ and the output is a predicted probability for $Y_0=1$. We describe how $H$ is mapped to $X$ in Section \ref{mapping}.

Let $L$ be the number of hidden layers in the FCNN. Let $l=0$ and $l=L+1$ correspond to the input and output layers respectively. Let $N_l$ be the number of nodes in layer $l$, where $N_0$ is the length of $X$ and $N_{L+1}=1$. The outputs of the layers are 
\begin{flalign*}
a^0 &= X \in \mathds{R}^{N_0}, \\
a^l &= \phi^l(W^l a^{l-1} + b^l) \in \mathds{R}^{N_l},  \qquad (l=1,\dots,L),
\end{flalign*}
where $W^l \in \mathds{R}^{N_l * N_{l-1}}$ is the matrix of weights for layer $l$ with row $i$ containing the weights of node $i$, $b^l \in \mathds{R}^{N_l}$ is the bias vector for layer $l$, and $\phi^l:\mathds{R}^{N_l} \to \mathds{R}^{N_l}$ represents the component-wise application of an activation function  $\phi:\mathds{R}\to\mathds{R}$. Commonly used activation function include the logistic function $\sigma(z) = \frac{\exp(z)}{1+\exp(z)}$ and the rectifier function $ReLU(z) = \max(0, z)$. The output layer ($l=L+1$) consists of a single node that uses the logistic activation function, outputting the predicted probability 
\[ \hat Y_0 = a^{L+1} = \sigma({W^{L+1}} a^L + b^{L+1}).\]

Given a cost function $C$ and $M$ training observations $(X_m, {Y_0}_m)$, $m=1,\dots,M$, the weight and bias parameters are randomly initialized and iteratively updated to minimize $\sum_{m=1}^{M} C({Y_0}_m, \hat{Y_0}_m)$ using methods such as stochastic gradient descent \citep{kiefer1952stochastic} and the Adam optimizer \citep{kingma2014adam}. Examples of cost functions \citep{janocha2017loss} include mean squared error, $C(y, z) = (y-z)^2$, and cross-entropy loss, $C(y, z) = -y \log(z) - (1-y)\log(1-z)$. When squared error loss or cross-entropy loss is used, then it is appropriate to interpret the NN output as a probability \citep{hampshire1991equivalence}.

The number of parameters ($W,b$) in a FCNN grows quickly with the size of the input and the number and size of the hidden layers. Various regularization methods have been developed to avoid overfitting, such as dropout \citep{srivastava2014dropout}. 

\subsection{Standardizing and Flattening Pedigrees}\label{mapping}

Since FCNNs require a fixed-size input, they cannot be directly applied to pedigrees, which vary in size and structure. It is possible to generate a fixed-size input based on simple summaries of family history, but this can result in substantial loss of information. Therefore, we propose the following approach: define a reference pedigree with pre-specified relatives (for example: counselee, grandparents, parents, sister, brother) and map each actual pedigree $H$ to a standardized version $H'$ that matches the structure of the reference pedigree (each relative in the reference pedigree may or may not be present in the actual pedigree), then flatten $H'$ into a fixed-length vector input $X$ for a FCNN. 

\begin{figure}[h!]
\centering
  \includegraphics[width=\columnwidth]{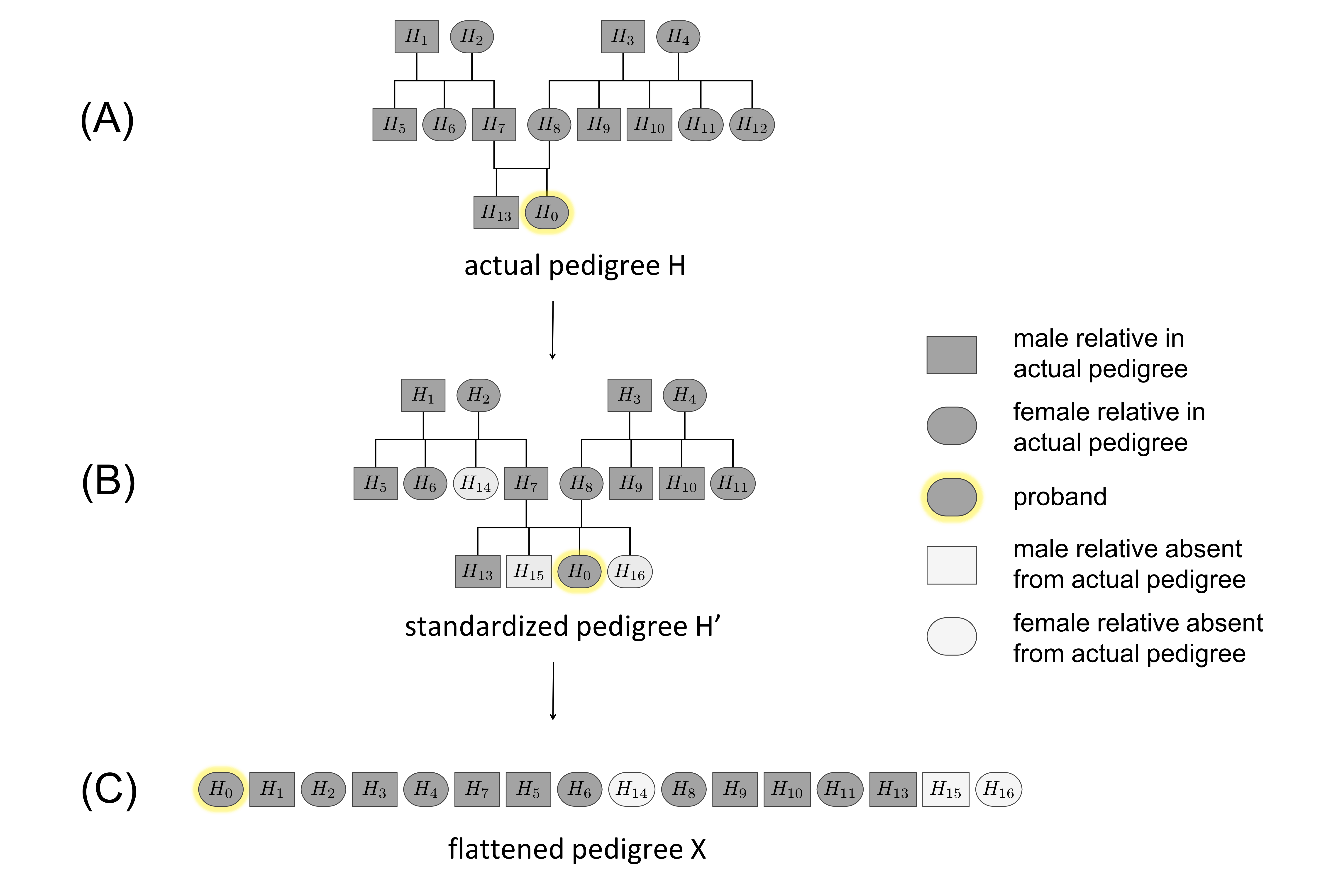}%
  \caption{Consider a reference pedigree that includes the counselee's grandparents, parents, uncles, aunts, and siblings, with each couple having 2 children of each sex. (A) Actual pedigree $H$. (B) Standardized pedigree $H'$, obtained by mapping $H$ to the reference structure. The actual pedigree has more maternal aunts than the reference pedigree, so we randomly select the desired number of maternal aunts to include in $H'$. The actual pedigree has fewer paternal aunts, sisters, and brothers than the reference pedigree, so in $H'$, we use pre-specified non-informative values for the paternal aunts, sisters, and brothers absent from $H$. (C) Flattened pedigree $X$, which is used as input for a FCNN.}
  \label{fig:mapped}
\end{figure}

We first describe the reference pedigree (see Figure \ref{fig:mapped}(B) for an example of a reference structure). Let the reference pedigree contain the counselee and $Q'$ other types of relatives (mother, father, sister, brother, etc). Let $q=0, 1, \dots, Q'$ index the relative types, with $q=0$ corresponding to the counselee. Let $R_q'$ be the number of relatives of type $q$ for $q \in \{0, 1, \dots, Q'\}$. Let the family members be indexed by $r=0,1,\dots,R'$, where $R'=\sum_{q=1}^{Q'} R_q'$, $r=0$ corresponds to the counselee, $r=1,\dots,R_1'$ correspond to relatives of type $q=1$, $r=R_1'+1, \dots, R_1'+R_2'$ correspond to relatives of type $q=2$, and so on. 

The choice of the reference structure should depend on the family structures observed in the training data, and it is a compromise between model complexity / computational costs and potential loss of information. Since every counselee has two parents and four grandparents, the reference structure should at least include these relatives (assuming that most counselees provide information on these relatives). For other relatives, one approach is to calculate a summary measure, such as the median, for the number of relatives of each type (example: sister, brother, etc.) in the training data and define a reference structure where the number of relatives of a given type is equal to the value of the summary measure for the number of relatives of that type (example: if the median number of sisters is one in the training data, then include one sister in the reference structure). In order to reduce potential loss of information, the median can be replaced with a higher threshold, such as the third quartile. The amount of information lost can be quantified for each reference structure using the mean proportion of family members dropped from the original pedigree. The reference structure can then be chosen based on the investigator’s judgment of how much information loss is acceptable (this can be informed by prior knowledge or a sensitivity analysis looking at performance metrics for models trained using different reference structures). Implementation details are provided in Section \ref{implementation}.

Now we consider an actual pedigree matrix $H$ and describe how to standardize and flatten it (Figure \ref{fig:mapped}). For $q=0, 1, \dots, Q'$, let $R_q$ be the number of relatives of type $q$ in $H$ ($R_0=1$). To construct a standardized pedigree matrix, $H'$, with the same structure as the reference pedigree matrix, we compare the number of relatives of type $q$ in the actual pedigree to the number in the reference pedigree for each $q \in \{0, 1, \dots, Q'\}$. If the two numbers are the same ($R_q = R_q'$), then we include all of the $R_q'$ actual relatives in $H'$. If the actual number is smaller than the reference number ($R_q < R_q'$), then we include the $R_q$ actual relatives in $H'$ and represent each of the $R_q'-R_q$ absent relatives using a vector of pre-specified null values (zeros). If the actual number is larger than the reference number ($R_q > R_q'$), then we randomly select $R_q'$ of the actual relatives to include in $H'$. We also include a column in $H'$ to indicate whether each row corresponds to a relative who is absent from the actual pedigree (0 if present, 1 if absent). Therefore, $H'$ is an $R'+1$ by $K+1$ matrix where each row consists of a family member's $K$ cancer history features, along with the presence/absence indicator. Let $H_r'$ be the vector for relative $r$ in $H'$. We flatten $H'$ by concatenating its rows to get a vector, $X=({H_0'}, {H_1'}, \dots, {H_{R'}'}) \in \mathds{R}^{(R'+1)*(K+1)}$, which can be used as input to a FCNN. If there are additional features of interest beyond the family history features specified above (e.g., breast density), then they can simply be appended to the input vector $X$.

\subsection{Convolutional Neural Networks}\label{pedigreeCNNs}

FCNNs are prone to overfitting since the number of parameters grows quickly with network size \citep{geman1992neural}. CNNs \citep{lecun1998gradient}, which are widely used in problems where the input has a spatial structure, such as image classification, reduce the number of parameters by using convolutional layers that enforce selective connections and weight sharing. A convolutional layer can be viewed as a fully-connected layer where certain weights are set to 0 and certain weights are constrained to have the same value. To exploit the correlation structure of the input (for example, pixels that are spatially close often have highly correlated values), a convolutional layer applies the same functions (e.g. $x \rightarrow \max(0, wx)$) repeatedly to different fixed-size neighborhoods of the input (for example, sets of neighboring pixels). These functions are called convolutional filters. The number of parameters in these local functions depends on the choice of reference pedigree and on $K$, the number of features considered for each family member. The reference pedigree and $K$ can vary across different applications (for example, the available family history information might be more detailed in some datasets than others), and therefore the corresponding local functions are tailored and  applied to domains with distinct dimensionalities.

Analogous to neighboring pixels, closely related individuals are likely to have similar levels of susceptibility to cancer due to genetic similarity and shared environment. Therefore, we propose to adapt CNNs to pedigree data. For reference, a description of a standard CNN is provided in SM \ref{appendix:cnn}. While standard CNNs were designed for inputs that have a fixed size and structure, various generalizations have been proposed for graphs that vary in size and structure \citep{wu2019comprehensive,niepert2016learning}, such as molecular compounds. We follow  two main steps: 1) standardize the graphs to have the same size and structure, then 2) define a sequence of neighborhoods within each standardized graph and apply convolutional filters to those neighborhoods.  

Our approach leverages the structure of pedigrees.
Like in the FCNN approach, we use a standardized and flattened pedigree $X$ as the input (Figure \ref{fig:mapped}). Prior to running the CNN, for each family member $r$ in $H'$, we define a fixed-size neighborhood centered at $r$ consisting of $r$ and $r$'s first-degree relatives: self, mother, father, $m_1$ sisters, $m_2$ brothers, $m_3$ daughters, and $m_4$ sons. Similar to Figure \ref{fig:mapped}, if $r$ has more than $m_1$ sisters, then $m_1$ of them are randomly selected, and if $r$ has fewer than $m_1$ sisters, then we use a pre-specified index representing an absent relative whose features are set to zero (analogous to zero padding in standard CNNs, as described in SM \ref{appendix:cnn}). The same approach is used for brothers, daughters, and sons. The neighborhood is represented by a vector $\mathcal{N}(r)$ of length $U=3 + \sum_{i=1}^4 m_i$. Within $\mathcal{N}(r)$, the individuals are ordered by relative type with respect to $r$.

\begin{figure}[h!]
  \includegraphics[width=\columnwidth]{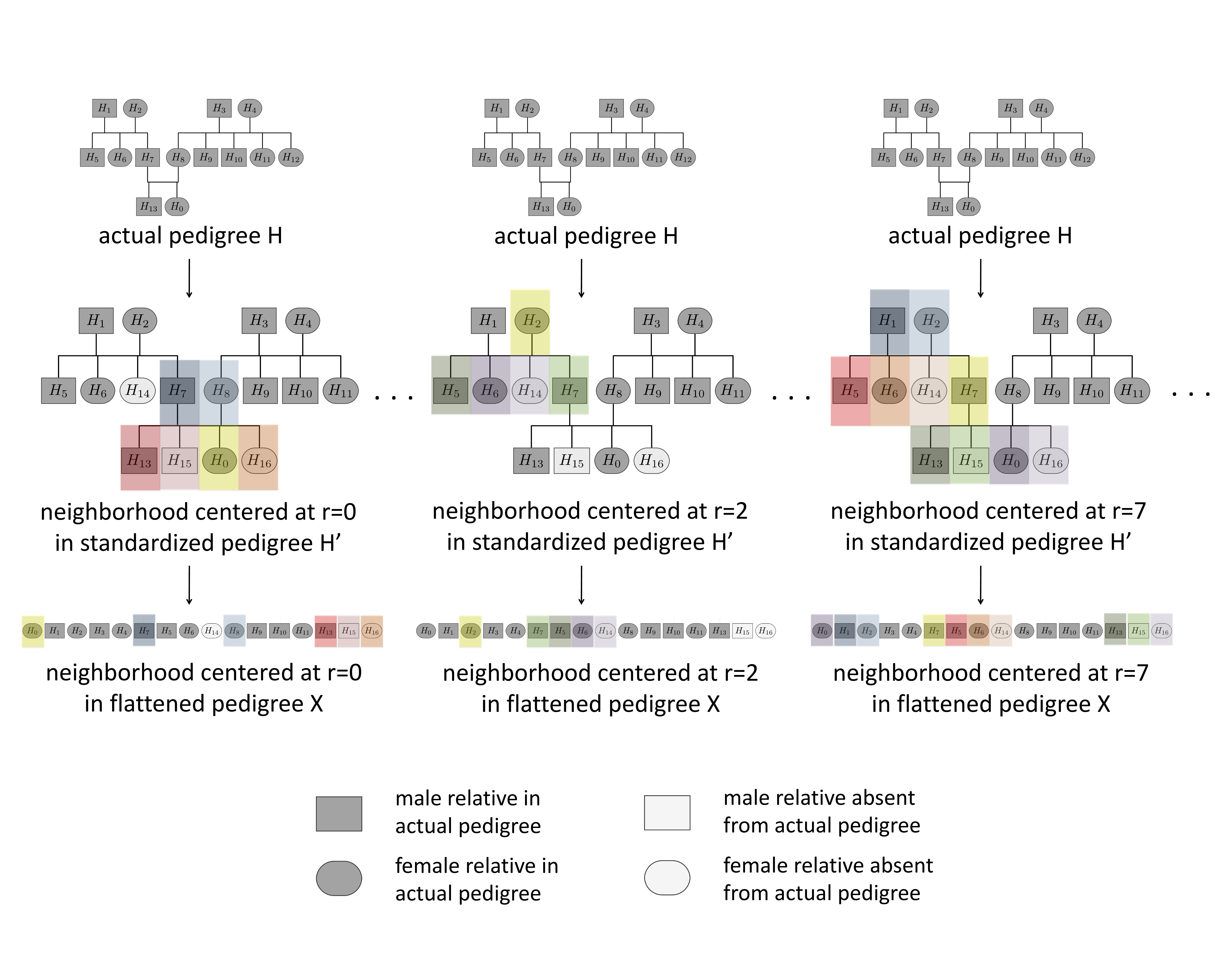}%
  \caption{The neighborhoods centered at relatives 0, 2, and 7 are shown above using shaded boxes. The same convolutional filters are applied to all neighborhoods of the pedigree. }
  \label{fig:cnn}
\end{figure}

We propose a CNN where all of the hidden layers are convolutional. There are $L$ hidden layers. Hidden layer $l$ applies $M_l$ real-valued convolutional filters $f_1^l, \dots, f_{M_l}^l$ to each of the $R'+1$ neighborhoods of the pedigree (Figure \ref{fig:cnn}). For $i=1,\dots,M_l$, let $f_i^l: \mathds{R}^{U * {M_{l-1}}} \to \mathds{R}$ (let $M_0=K+1$ since each relative has $K+1$ features in $H'$ - see Section \ref{mapping}). Let $a_{r}^l \in \mathds{R}^{M_l}$ be the output of layer $l$ for neighborhood/family member $r$. Let $a_{\mathcal{N}(r)}^{l-1} \in \mathds{R}^{U * M_{l-1}}$ be the vector obtained by concatenating the layer inputs of the relatives in $\mathcal{N}(r)$. The output from applying filter $i$ to $r$'s neighborhood is 
\[f_i^l(a_{\mathcal{N}(r)}^{l-1}) = \phi \left( w_i^l \cdot a_{\mathcal{N}(r)}^{l-1} + b_i^l \right),\]
where $\cdot$ is the dot product, $w_i^l \in \mathds{R}^{U * M_{l-1}}$ is the vector of weights for filter $i$ and $b_i^l \in \mathds{R}$ is the bias for filter $i$.

Let $f^l = (f_1^l, \dots, f_{M_l}^l): \mathds{R}^{U * M_{l-1}} \to \mathds{R}^{M_l}$. The layer outputs for relative $r$ are
\vspace{-0.25cm}
\begin{flalign*}
a_r^0 &= H_r' \in \mathds{R}^{K+1}, \;\;\;\;\;\text { for  }  l=0,  \text{ and }\\
a_r^l &= f^l(a_{\mathcal{N}(r)}^{l-1}) \in \mathds{R}^{M_l},\;\;\;\;\; (l=1, \dots, L),
\end{flalign*}
\vspace{-0.25cm}
and the overall layer outputs are 
\begin{flalign*}
a^l &= \left( {a_0^l} \, , \, \dots \, , \, {a_{R'}^l} \right) \in \mathds{R}^{M_l * (R'+1)}, \;\;\;\;\; (l=0,1, \dots, L).
\end{flalign*}

The final output is a transformation of $a_0^{L}$ using a logistic activation function: 
\[ \hat Y_0 = \sigma({w^{L+1}} \cdot a_0^L + b^{L+1})\]
where $w^{L+1} \in \mathds{R}^{M_L}$ and $b^{L+1} \in \mathds{R}$.

As in FCNNs, the weight and bias parameters are optimized with respect to $\sum_{m=1}^{M} C({Y_0}_m, \hat{Y_0}_m)$ and the optimization can be carried out using stochastic gradient descent.

There are various ways to incorporate additional features beyond family history. Additional features that are applicable to all relatives (e.g. body mass index) can be appended to the input vector $a_r^0$ for each relative $r$. For additional features that are only applicable to the counselee, a modification to the appendment approach is necessary since the use of convolutional filters requires the input vector for each relative to have the same size. Two possible approaches are: 1) append the features to the input vector of each relative, but set their values to 0 for non-counselees or 2) append the features for the counselee to the output vector of the $L$-th convolutional layer (i.e. the layer before the final output layer),  thus expanding  the input vector for the final output layer (a related approach is used in \citep{li2017concatenating}).

\subsubsection{Model Space}

Universal approximation theorems characterize the approximation capabilities of models and algorithms. The universal approximation theorem for FCNNs indicates that any continuous function over a given domain (e.g., the real line) can be approximated with arbitrary precision by a FCNN with a single hidden layer \citep{cybenko1989approximation, hornik1991approximation, leshno1993multilayer}. The theorem establishes the existence of a FCNN that satisfies the desired level of precision but does not provide a practical way to construct it.  In our setting, this is an attractive property because it means that any continuous relation between  the family history (in the form of a fixed-length vector) and cancer risk can be approximated  arbitrarily well by a FCNN. We show in this section that the CNNs we propose are just as powerful: they satisfy a universal approximation property similar to that of FCNNs.

Fix a reference pedigree $H^*$ of size $R'+1$ containing relatives of up to degree $d$ of the counselee. Let $Q'$ be the number of relative types in $H^*$ besides the counselee and let $m = \max\limits_{q=0,1,\dots,Q'} R_q'$. Let $\mathcal{X}^* \subset \mathds{R}^{(R'+1)*(K+1)}$ be the space of pedigrees with the same structure as $H^*$. We consider the CNN's ability to approximate functions from $\mathcal{X}^*$ to $[0, 1]$. We first state the universal approximation theorem for standard FCNNs \citep{leshno1993multilayer} and then verify that the same property extends to CNNs (proof provided in SM \ref{appendix:proof}). 

\paragraph{Universal Approximation Theorem for FCNN} (forward direction of Theorem 1 from \citep{leshno1993multilayer})
Let $k$ be a positive integer and $I$ a compact subset of $\mathds{R}^k$. Let $g:I \to \mathds{R}$ be continuous. Let $\phi: \mathds{R} \to \mathds{R}$ be a piecewise continuous, locally bounded, and non-polynomial activation function. Then given $\epsilon>0$, there exists a positive integer $N$, and for $i=1,\dots,N$, constants $\alpha_i, b_i \in \mathds{R}$ and vectors $w_i \in \mathds{R}^k$ such that
\[ F(X) = \sum_{i=1}^N \alpha_i \phi(w_i \cdot X + b_i), \]
satisfies $|F(X) - g(X)|<\epsilon$ $\forall X \in I$. 
\par

\begin{theorem}\label{cnn_thm} [Universal Approximation Theorem for Pedigree CNNs] 
Assume that the elements of $H_r' \in \mathds{R}^{K+1}$ are bounded for $r=0,1,\dots,R'$. Let $g:\mathcal{X}^* \to [0,1]$ be continuous. Let $\phi: \mathds{R} \to \mathds{R}$ be a continuous and invertible activation function. Let the fixed-size neighborhood about each relative contain $m_1=\dots=m_4=m$ sisters/brothers/daughters/sons. Then given $\epsilon>0$, there exists a pedigree CNN of the form described in Section \ref{pedigreeCNNs} with $d$ hidden layers with activation function $\phi$, $M_l$ convolutional filters for hidden layer $l$, bias terms $b_i^l \in \mathds{R}$ ($i=1,\dots,M_l$; $l=1,\dots,L+1$), and weight vectors $w_i^l \in \mathds{R}^{U * M_{l-1}}$ ($i=1,\dots,M_l$; $l=1,\dots,L+1$), such that the final output
\[ F(X) = \sigma(w^{L+1} \cdot a_0^{L}(X) + b^{L+1}) \]
satisfies 
$|F(X) - g(X)|<\epsilon$ $\forall X \in \mathcal{X}^* $.
\end{theorem}

\subsection{Missing Data}

In practice, there is often missing information in family history data (for example, an unreported relative or an unknown diagnosis age). Missing values in the training and/or test set can be handled using standard imputation methods or complete case analysis \citep{little2019statistical}, though the latter may result in a substantial decrease in sample size. Missing value imputation can be implemented as a preprocessing step separate from training or prediction \citep{garcia2010pattern}. In clinical practice, some models do not allow missing values (e.g. the Claus model \citep{claus1994autosomal}),  and clinicians impute missing information (e.g. ages of diagnosis for relatives) to compute predictions. Some popular clinical tools  automatically impute  missing information. For example, in the Risk Service tool, a missing diagnosis age for a relative is imputed based on the relative’s current age \cite{chipman2013providing}.

Another approach that can be implemented for NNs and prediction models in general is to include as predictors indicator functions denoting whether certain features are missing \cite{choi2019comparison}. In our analyses, we used this approach to represent absent family members when mapping families to a reference pedigree (Figure \ref{fig:mapped}) that potentially contains relative types absent from the actual family. Since missing values are distinct from nonexistent data, separate indicators could be used for missingness versus absence.

As described in Section \ref{sensitivity}, we performed a sensitivity analysis using simulated data to evaluate the impact of missing relatives and missing ages of diagnosis.

\subsection{Benchmark Methods}

In our simulations and data application, we focused on breast cancer risk prediction and compared NNs to the Mendelian BRCAPRO model and to LR, which is equivalent to a single-node FCNN with a logistic activation function. For LR, we used the flattened pedigree $X$ as the input.

BRCAPRO \citep{berry1997probability, parmigiani1998determining} is widely used in clinical practice and has been validated in various populations \citep{berry2002brcapro, euhus2002pretest, terry201910, mccarthy2019performance}. It estimates the probability of carrying a germline mutation in breast/ovarian cancer susceptibility genes BRCA1 and BRCA2, as well as future risk of breast/ovarian cancer, using Bayes' rule, laws of Mendelian inheritance, mutation prevalence and penetrance, and family history of breast and ovarian cancer. The family history information includes the $K=6$ features described in Section \ref{notation}: breast/ovarian cancer status, age at onset of breast/ovarian cancer if applicable, and current age or age at death. In addition, BRCAPRO provides the option of modifying the default prevalences and penetrances using the following covariates, if they are available: race, ethnicity, genetic testing results for BRCA1/BRCA2, marker testing results (ER/CK14/CK5/CK6/PR/HER2), and prophylactic mastectomy/oophorectomy (these additional covariates were not included in our simulations). 

Let $\gamma_r$ be the genotype of relative $r$ (non-carrier, carrier of a pathogenic BRCA1 mutation, carrier of a pathogenic BRCA2 mutation, or carrier of pathogenic mutations in both BRCA1 and BRCA2). Using Bayes' rule and the assumption of conditional independence of phenotypes given genotypes, the counselee's probability of having genotype $\gamma_0$ is
\begin{equation}\label{eq:carrier}
P(\gamma_0|H) = \frac{P(\gamma_0) \sum_{\gamma_1, \dots, \gamma_R} \prod_{r=0}^R P(H_r|\gamma_r)  P(\gamma_1, \dots, \gamma_R|\gamma_0) }{\sum_{\gamma_0} P(\gamma_0) \sum_{\gamma_1, \dots, \gamma_R} \prod_{r=0}^R P(H_r|\gamma_r)  P(\gamma_1, \dots, \gamma_R|\gamma_0) }.
\end{equation}
The summation over genotypes is calculated using the Elston-Stewart peeling algorithm \cite{elston1971general} and $P(\gamma_1,\dots, \gamma_R|\gamma_0)$ is calculated based on Mendelian laws of inheritance. The prevalences $P(\gamma_r)$ are obtained from the literature and are ethnicity-specific (in particular, different prevalences are used for Ashkenazi Jewish and non-Ashkenazi Jewish families). $P(H_r|\gamma_r)$ is calculated using literature-based penetrances for breast and ovarian cancer. The penetrances are functions that represent  the risk of  cancer at  different  ages  age and  they are genotype- cancer- and sex-specific. The penetrance functions for non-carriers are based on rates from the Surveillance, Epidemiology, and End Results (SEER) program and are race-specific, while the penetrance functions for carriers are from a meta-analysis of published studies \cite{chen2020penetrance}.

After estimating the carrier probabilities, BRCAPRO calculates future risk of breast cancer through a weighted average of the genotype-specific penetrance functions $P(Y_0=1|\gamma_0)$:
\[ P(Y_0=1|H) = \sum_{\gamma_0} P(Y_0=1|\gamma_0) P(\gamma_0|H_0, \dots, H_R). \]

\subsection{Model Evaluation}

We evaluated model performance using four metrics \citep{steyerberg2010assessing}: 1) the ratio of observed (O) to expected (E) events (where E is the sum of the predictions in the test set), a measure of calibration, 2) the area under the receiver operating characteristic curve (AUC), a measure of discrimination, 3) the area under the precision recall curve (PR-AUC), another measure of discrimination that is more sensitive to class imbalance than the AUC, and 4) the Brier score, which is the mean squared difference between the predicted probabilities and actual outcomes. We obtained 95\% confidence intervals (CIs) for the metrics by bootstrapping the test set 1000 times.

\subsection{Implementation}\label{implementation}

We ran BRCAPRO using the BayesMendel R package (version 2.1-6) \citep{BayesMendel}. The NNs were implemented in Python using Keras (\url{https://github.com/keras-team/keras}) with the Theano backend \citep{team2016theano}. For the CNNs, we adapted code from \cite{hechtlinger2017generalization}. 

In the simulations, 887,353 randomly generated families were split into a training set of 800,000 and a test set of 87,353. In the data application, the Risk Service dataset (279,460 families) was used for training and the CGN dataset (7489 families) was used for testing. In both the simulations and data application, we used the Adam optimizer \cite{kingma2014adam} and the mean squared error loss function (while cross-entropy loss is more commonly used for binary outcomes, we chose to use mean squared error because it corresponds to the minimization of the Brier score, which is a standard performance metric in risk prediction \cite{steyerberg2010assessing} and one of the metrics we used to compare models; more discussion on this choice and a sensitivity analysis are provided in SM \ref{appendix:loss}). We used a typical 90/10 split of the training set to tune NN hyperparameters via a random search \cite{bergstra2012random}: 10\% of the training set was held out for evaluating the performance of different choices for the number of hidden layers (1 to 3), sizes of hidden layers (10 to 100), number of filters for the CNN (3 to 10), learning rate (0.0001 to 0.01), weight decay parameter (0 to 0.01), activation function (ReLU, or elu), and dropout rate (0 to 0.5). The performance in the tuning set was highly sensitive to the hyperparameter values (in the simulations, AUCs in the held out subset ranged from 0.38-0.65 for the FCNN and 0.56-0.65 for the CNN: https://github.com/zoeguan/nn\_cancer\_risk/tree/master/tuning\_results), so it is important to explore different sets of hyperparameters.

In the simulations, the FCNNs had 2 hidden layers of sizes 30 and 10 and the CNNs had 2 convolutional layers with 10 and 5 filters. In the data application, the FCNN had 2 hidden layers of size 30 and the CNN had 2 convolutional layers with 5 filters each. We also used a dropout layer following the first hidden layer in each NN, with a dropout rate of 20\%. We used the Exponential Linear Unit (ELU) activation function \citep{clevert2015fast}. For the NNs and LR, features were normalized to be between 0 and 1 using min-max normalization \citep{patro2015normalization}. The code for the analyses is available at \url{github.com/zoeguan/nn_cancer_risk} and contains additional details on hyperparameter values.

In the simulations, we used a reference pedigree of size 26 containing the counselee's grandparents, parents, aunts (2 maternal, 3 paternal), uncles (3 maternal, 2 paternal), siblings (2 sisters, 3 brothers), and children (2 daughters, 2 sons). This was chosen based on the distribution of family structures in the CGN (see SM \ref{appendix:refped}). We used $m_1=m_2=3$ and $m_3=m_4=2$ for the CNN neighborhoods. In the data application, we used a reference pedigree of size 19 with the same relative types as in the simulations, but restricted to 2 relatives of each type and omitted sons and daughters due to the smaller family sizes in the training dataset (see SM \ref{appendix:refped}). We used $m_1=m_2=2$ and $m_3=m_4=1$ for the CNN neighborhoods. 

Studies have shown that restricting family history to first- and second-degree relatives \cite{biswas2013simplifying,terry201910} has little impact on discriminative accuracy. Therefore we considered only first- and second-degree relatives in the reference pedigree. As described in Section \ref{sensitivity}, we conducted a sensitivity analysis for various choices of reference pedigree structures and found little variation in performance.

\section{Simulations}

We evaluated the performance of the proposed NN approaches in predicting 10-year risk of breast cancer in two simulation settings: one where the data are consistent with BRCAPRO and one where they are not.

\subsection{Simulation Approach}
We simulated 1,000,000 pedigrees using the generating model assumed by BRCAPRO. To simulate each family, we first sampled a family structure (number of sisters, brothers, etc) from the CGN dataset (described in Section \ref{datasets}). For counselees, we also sampled dates of birth and baseline dates for risk assessment from the CGN. For non-counselees, dates of birth were generated relative to the counselee's date of birth by assuming that the age difference between a parent and a child has mean 27 and standard deviation 6. 

Next, we generated the genotypes for each family member. We first generated the genotypes of the counselee's grandparents (the oldest generation) using the default Ashkenazi Jewish allele frequencies in BRCAPRO (0.014 for BRCA1 and 0.012 for BRCA2) to mimic a higher-risk population. For individuals in subsequent generations, we generated genotypes according to Mendelian inheritance. 

We generated ages of onset for breast and ovarian cancer conditional on the genotypes. Each age of onset was randomly generated from \{1, \dots, 94\}, with probabilities given by the genotype-specific penetrance functions from BRCAPRO (the cumulative lifetime probability of breast cancer ranges from 0.12 for non-carriers to 0.79 for carriers of mutations in both BRCA1 and BRCA2). We also generated a death age for each individual from a distribution with mean 80 and standard deviation 15. If an individual's age of onset was greater than their baseline age or death age, then their cancer status at baseline was set to 0.

\subsection{Results}\label{sim_results}

We excluded counselees who died or were diagnosed with breast cancer prior to baseline. For the remaining counselees ($n=887,353$), we predicted 10-year risk of breast cancer using the baseline family history. We used 800,000 families for training and the other 87,353 for testing. In the training set, there were 23,606 cases (counselees who developed breast cancer within 10 years). In the test set, there were 2570 cases.

\begin{figure}[h!]
\centering
  \includegraphics[width=0.9\columnwidth]{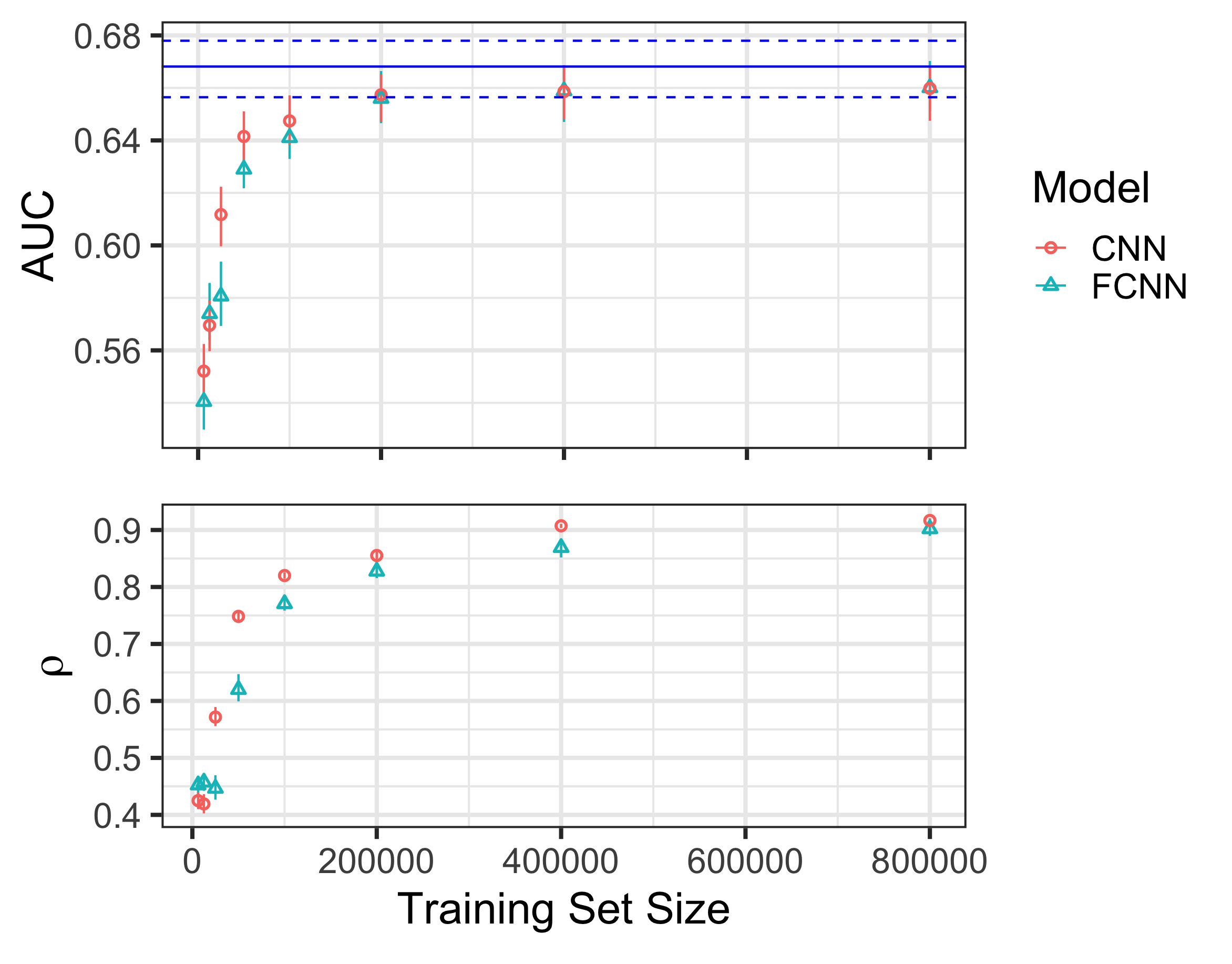}
  \caption{AUC and correlation ($\rho$) of NN predictions with BRCAPRO predictions for 10-year risk of developing breast cancer as a function of training sample size (ranging from to 6,250 to 800,000) in simulations.}
  \label{fig:auc}
\end{figure}

We investigated how much training data is needed for the performance of the NNs to approach that of the true model by training NNs on increasingly large subsets of the entire training set, with sample sizes ranging from 6,250 to 800,000 (Figure \ref{fig:auc}). As the sample size increased, the AUCs of the NNs approached that of BRCAPRO, the true data generating model, and the predictions from the NNs became highly correlated with those from BRCAPRO. For sample sizes under 100,000, the CNN had a higher AUC than the FCNN, though, as expected, the differences between the two approaches decreased with increasing sample size. With 200,000 or more training examples, both the FCNN and CNN achieved AUCs similar to that of the true model (both NNs had an AUC of 0.660 while the true model had an AUC of 0.668). 

\begin{figure}[htb!]
\centering
  \includegraphics[width=0.9\columnwidth]{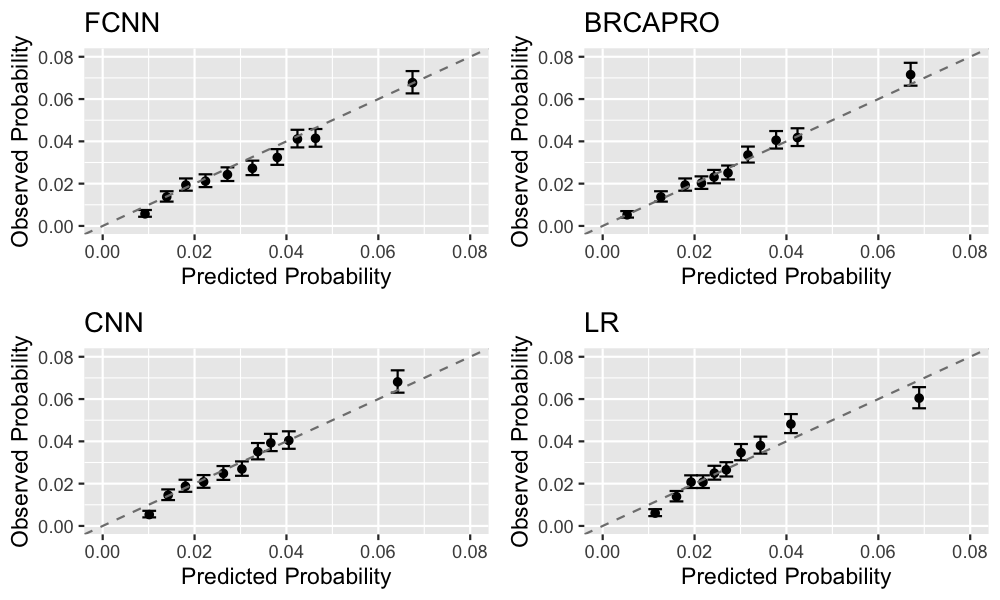}
  \caption{Calibration plots by decile of risk in simulated families (training set of 800,000).}
  \label{fig:cal}
\end{figure}

The NNs provided a better approximation of the true model than LR. The FCNN and CNN trained on the entire training set achieved correlations of 0.9 and 0.92 with the true model, while the LR model trained on the same data had a correlation of 0.82 (``True Family History" section in Table \ref{table:sim}). The NNs also outperformed LR with respect to AUC, PR-AUC, and Brier score: across 1000 bootstrap replicates of the test set, the NNs had a better AUC and Brier score than LR more than 99\% of the time. The proportion of cases in our dataset is very small, therefore all of the models have low PR-AUCs (the baseline PR-AUC, or PR-AUC of a model that does no better than random guessing, is the proportion of cases, 0.029). The CNN was more highly correlated with BRCAPRO than the FCNN across all 1000 bootstrap replicates. Also, the CNN had a better Brier score than the FCNN in more than 95\% of the bootstrap replicates and a higher AUC in 58\% of the replicates. The CNN and LR both had good overall calibration, with O/E=0.99 (95\% CI 0.95-1.03) for the CNN and O/E=1.00 (95\% CI 0.96-1.04) for LR (Table \ref{table:sim}), while the FCNN slightly overestimated risk, with O/E=0.93 (95\% CI 0.89-0.96). Across the bootstrap replicates, the CNN and LR performed similarly with respect to calibration, with the CNN showing better calibration in about half of the replicates. The CNN and LR had better calibration than the FCNN in more than 97\% of the replicates. Calibration plots by decile of estimated risk (Figure \ref{fig:cal}) show that LR underestimated or overestimated risk in more deciles compared to the other models. We also plotted the precision-recall curves for the models (Figure \ref{fig:pr_curve}), which were not substantially different across models.

\begin{figure}[h!]
\captionsetup{singlelinecheck=off}
\footnotesize
\centering
\includegraphics[width=\columnwidth]{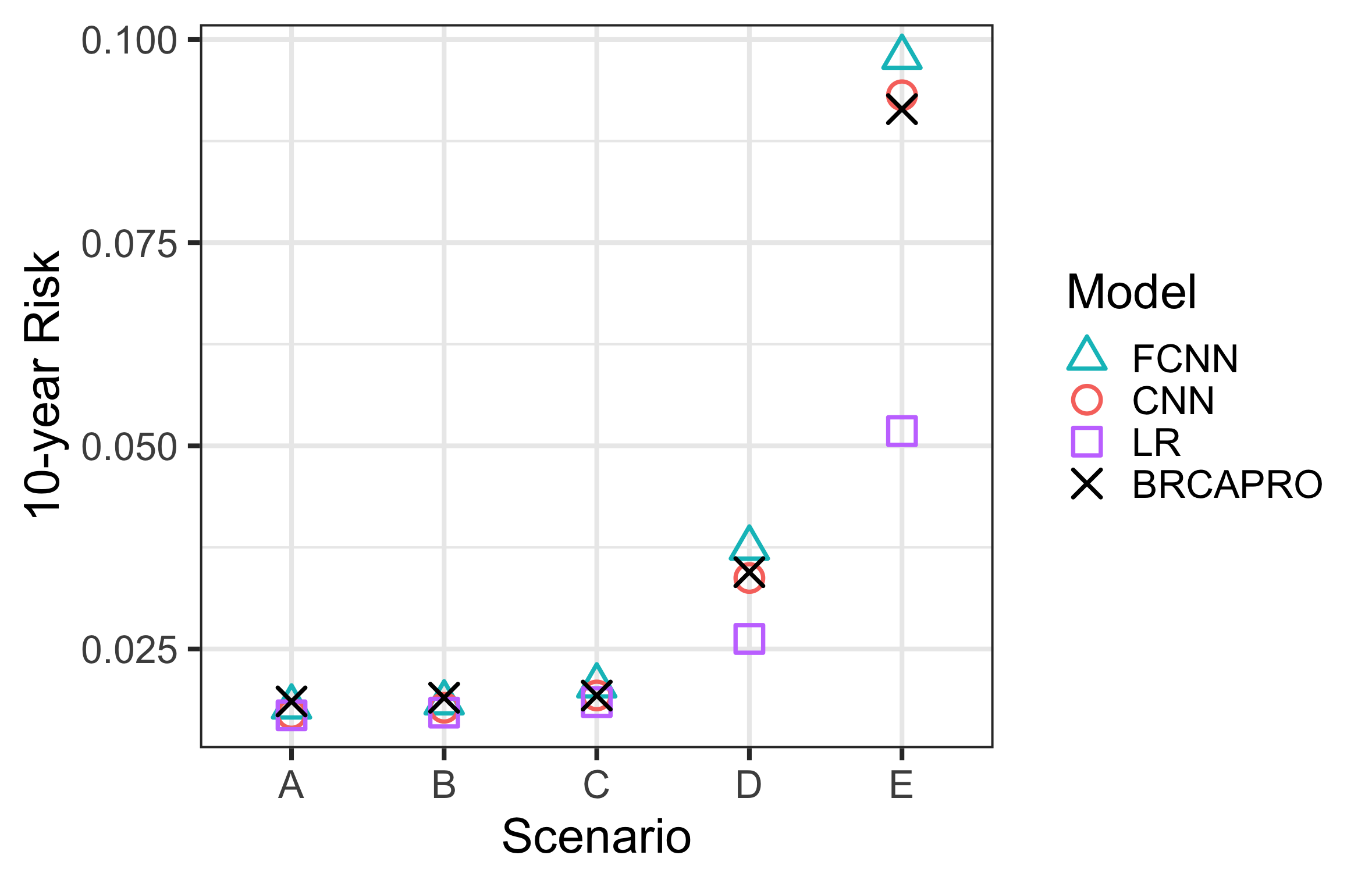}
\caption[]{Under the first simulation setting, we fixed a simulated family structure (counselee, grandparents, parents, 1 paternal aunt, 1 maternal aunt, 2 maternal uncles) for a 40-year-old counselee and varied the level of family history across 5 scenarios (ordered by increasing risk with respect to BRCAPRO, the true model):
\begin{enumerate}
\item no affected relatives
\item maternal grandmother diagnosed with breast cancer at age 80
\item maternal grandmother diagnosed with breast cancer at age 60
\item maternal grandmother diagnosed with breast cancer at age 60, mother diagnosed with breast cancer at age 50
\item maternal grandmother diagnosed with breast cancer at age 60, mother diagnosed with breast cancer at age 50 and ovarian cancer at age 60. We calculated 10-year risk predictions for each scenario using each model
\end{enumerate}}
\label{fig:family_interactions}
\end{figure}

{\bf  Differences between LR and CNN. } Under the true model, the counselee's risk of breast cancer increases with more affected relatives and earlier diagnosis ages. To assess whether NN and LR predictions captured these trends, we fixed a family structure and varied the phenotypes of the mother and maternal grandmother (Figure \ref{fig:family_interactions}). We considered five scenarios ordered by increasing risk with respect to the true model: (A) no affected relatives, (B) grandmother with breast cancer, (C) grandmother with breast cancer at an earlier age, (D) grandmother with breast cancer, mother with breast cancer, and (E) grandmother with breast cancer, mother with breast and ovarian cancer. While the NNs gave similar predictions to BRCAPRO across all scenarios (Figure \ref{fig:family_interactions}), LR slightly underestimated risk in Scenario (D) and severely underestimated risk in Scenario (E). LR assumes a restrictive functional form for the relationship between the features and the outcome, and this functional form does not match that of BRCAPRO, the data generating model, so the LR model is misspecified in these simulations. NNs with multiple hidden nodes are more flexible than LR and therefore less susceptible to misspecification.

\begin{table}[!htbp]
\centering
\begingroup\scriptsize
\begin{tabular}{llllll}
  \hline
 & O/E & $\Delta$AUC & $\Delta$PR-AUC & $\Delta$sqrt(BS) & $\rho$ \\ 
  \hline\underline{\textbf{True Family History}} & & \\ 
 \textbf{Performance Metrics} & & \\ 
FCNN & 0.93 (0.89, 0.96) & -1.21 (-1.73, -0.63) & -10.16 (-13.81, -7.08) & -0.19 (-0.26, -0.12) & 0.90 (0.89, 0.91) \\ 
  CNN & 0.99 (0.96, 1.03) & -1.24 (-1.80, -0.69) & -7.93 (-11.52, -4.35) & -0.14 (-0.19, -0.10) & 0.92 (0.91, 0.92) \\ 
  LR & 1.00 (0.97, 1.04) & -2.07 (-2.68, -1.47) & -14.59 (-19.04, -10.25) & -0.28 (-0.36, -0.21) & 0.82 (0.81, 0.83) \\
  BRCAPRO & 1.02 (0.98, 1.06) & AUC=0.668 & PR-AUC=0.065 & sqrt(BS)=0.168 & 1.00 (1.00, 1.00) \\ 
   \multicolumn{3}{l}{\textbf{Comparisons Across Bootstrap Replicates}} \\ 
FCNN$>$CNN & 0.021 & 0.582 & 0.020 & 0.038 & 0.000 \\ 
  FCNN$>$LR & 0.025 & 1.000 & 0.990 & 0.991 & 1.000 \\ 
  FCNN$>$BRCAPRO & 0.083 & 0.000 & 0.000 & 0.000 & 0.000 \\ 
  CNN$>$LR & 0.464 & 1.000 & 0.999 & 1.000 & 1.000 \\ 
  CNN$>$BRCAPRO & 0.691 & 0.000 & 0.000 & 0.000 & 0.000 \\ 
   \hline\underline{\textbf{Misreported Family History}} & & \\ 
 \textbf{Performance Metrics} & & \\ 
FCNN & 1.06 (1.02, 1.10) & 2.82 (1.72, 3.99) & 9.31 (2.66, 16.43) & 0.48 (0.35, 0.60) & \\ 
  CNN & 1.01 (0.97, 1.05) & 2.70 (1.63, 3.72) & 11.15 (5.41, 17.47) & 0.54 (0.44, 0.64) & \\ 
  LR & 1.00 (0.96, 1.04) & 2.35 (1.23, 3.49) & 6.12 (0.41, 12.47) & 0.48 (0.37, 0.59) & \\
  BRCAPRO & 0.81 (0.78, 0.84) & AUC=0.627 & PR-AUC=0.050 & sqrt(BS)=0.169 &  \\ 
   \multicolumn{3}{l}{\textbf{Comparisons Across Bootstrap Replicates}} \\ 
FCNN$>$CNN & 0.033 & 0.666 & 0.232 & 0.038 &  \\ 
  FCNN$>$LR & 0.061 & 0.968 & 0.869 & 0.487 &  \\ 
  FCNN$>$BRCAPRO & 1.000 & 1.000 & 0.998 & 1.000 &  \\ 
  CNN$>$LR & 0.406 & 0.900 & 0.991 & 0.996 &  \\ 
  CNN$>$BRCAPRO & 1.000 & 1.000 & 1.000 & 1.000 &  \\ 
   \hline
\end{tabular}
\endgroup
\caption{Model performance in simulated families (training set of 800,000) based on true and misreported family history. $\Delta$AUC: \% relative improvement in AUC compared to BRCAPRO. $\Delta$PR-AUC: \% relative improvement in precision-recall AUC compared to BRCAPRO. $\Delta$sqrt(BS): \% relative improvement in root Brier Score compared to BRCAPRO. $\rho$: correlation with BRCAPRO. The ``Comparisons Across Bootstrap Replicates" section shows pairwise comparisons between the NN models and the other models across 1000 bootstrap replicates of the test set; the row for $A>B$ shows the proportion of bootstrap replicates where model A outperformed model B with respect to each metric. } 
\label{table:sim}
\end{table}

\subsubsection{Perturbations of Mendelian Models}

Misreported cancer diagnoses can considerably distort predictions from Mendelian models \citep{katki2006effect, braun2014extending}. In the second simulation setting, we introduced noise to the simulated family histories through incorrectly reported diagnoses, diagnosis ages, and current ages for non-counselees using misreporting rates from \cite{ziogas2003validation} and \cite{braun2017nonparametric} (see SM \ref{appendix:misreporting} for details).

Under misreporting, the NNs outperformed BRCAPRO with respect to calibration, AUC, and Brier score across almost all of the 1000 bootstrap replicates of the test set  (Table \ref{table:sim}), illustrating the advantage of NNs over BRCAPRO when the Mendelian assumptions are not fully satisfied. The NNs also outperformed LR with respect to AUC and PR-AUC in most of the bootstrap replicates. With respect to the Brier score, the CNN outperformed LR in more than 99\% of the replicates, while the FCNN performed similarly to LR. The CNN had similar calibration to LR, while the FCNN had worse calibration.

\subsection{Sensitivity Analyses}\label{sensitivity}

Using simulated data, we performed sensitivity analyses to evaluate the impact of the choice of reference pedigree and missing data. 

To evaluate the impact of the choice of reference pedigree, we quantified the amount of information lost (mean proportion of family members dropped from the original pedigree) for various reference structures based on different summary measures for the number of relatives of each type (SM \ref{appendix:refped}). We considered ``symmetric" reference structures where the number of daughters is equal to the number of sons for each couple, as well as reference structures without this constraint. We assessed the discriminatory accuracy of the models trained using the various reference structures. The results show only small differences in performance for reference structures based on using the first, second, third, or fourth quartile as the summary measure, even though the mean proportion of family members dropped varies substantially across these choices (from 0 for the fourth quartile to approximately 0.4 for the first quartile). Therefore, in our application, the performance of the NNs is not particularly sensitive to the choice of reference structure. In the main analyses, we used a symmetric reference structure based on the third quartile of relative counts. In other settings where performance may be more sensitive to the choice of reference structure, it can be chosen based on cross-validation AUCs or other performance metrics.

We also considered the impact of different proportions of missing data in the training and test sets. We evaluated the impact of 1) missing relatives by removing relatives from the pedigree and 2) missing diagnosis ages for affected relatives. In the first scenario, we considered removing relatives at random, which corresponds to non-informative missingness, as well as, removing only unaffected relatives, which corresponds to informative missingness. In prediction problems, missing data is likely to have a larger impact when the amount of missingness differs between the training and test datasets, so we considered scenarios where there were missing data in the training set but complete data in the test set, as well as symmetric scenarios with complete data in the training set and missing data in the test set. We varied the proportion of missing relatives and missing diagnosis ages from 0.05 to 0.3. We used single imputation to handle the missing ages, setting them to 50 for individuals over 50 and setting them to the individual’s current age otherwise. The types of missingness considered did not have a substantial impact on any of the performance measures (Tables \ref{table:missing_train}-\ref{table:missing_test2}).

\subsubsection{Computational Costs}

Among the models trained, LR is the least computationally intensive and CNN the most computationally intensive. The training times using a single CPU core for different training set sizes ranging from 6,250 to 800000 are provided in Figure \ref{fig:training_time} (using the NN hyperparameters from the main simulation analysis). The relationship between sample size and training time is approximately linear for each model. With 800,000 training families, it took about 1 minute to train the LR model, 5 minutes to train the FCNN, and 20 minutes to train the CNN. The NNs also require additional computation to tune the hyperparameters prior to training the final model, which can considerably increase the computational burden. However, hyperparameter tuning methods such as grid search and random search can be parallelized, and in some cases, using a GPU \cite{oh2004gpu} can speed up the tuning/training process.

\section{Data Application} \label{data}

We trained NN and LR models to predict 5-year risk of breast cancer using data from the Risk Service and compared their performance to BRCAPRO using data from the CGN. We excluded male counselees, counselees who had breast cancer/bilateral mastectomy/bilateral oophorectomy before baseline, counselees under 18 years old, and counselees for whom we could not run BRCAPRO (counselees over 89 years old).

\subsection{Datasets} \label{datasets}

\subsubsection{Risk Service}
The Risk Service \citep{chipman2013providing} is a web service that provides risk predictions from various family history-based cancer risk models, including BRCAPRO. It has been used in primary care, breast imaging, and genetic counseling clinics. As of January 2018, the Risk Service database contained patient-reported family history inputs for over 450,000 counselees, with 285,161 counselees consenting to the use of their data for research. 

Model training requires baseline and follow-up data, but the Risk Service does not follow counselees over time. We therefore defined each counselee's baseline date to be 5 years prior to the date at which they used the Risk Service and the follow-up date to be the date at which they used the Risk Service. We retrospectively reconstructed the family history at the baseline date based on the ages and diagnosis ages of the family members. However due to a considerable amount of missing age information for non-counselees (74\% of first- and second-degree relatives were missing age and 34\% of affected first- and second-degree relatives were missing age at diagnosis), we decided not to use ages or diagnosis ages of non-counselees for training and we imputed baseline cancer status for non-counselees with missing diagnosis ages (see SM \ref{appendix:missingages} for more details). 

The training set consisted of 279,460 counselees (Table \ref{cohorts}). The median age was 45 and the median family size was 8. Also, 36,783 counselees (13.2\%) had at least one affected first-degree relative and 13,307 (4.8\%) developed breast cancer during the follow-up period.

\subsubsection{CGN}
The CGN is a national consortium of 15 academic medical centers that was established for the purpose of studying inherited predisposition to cancer \citep{anton2003cancer}. Between 1999 and 2010, 26,941 participants with cancer or a family history of cancer were recruited through population-based registries, high-risk clinics, and self-referral. They provided information on personal and family history of cancer and sociodemographic factors through a baseline phone interview and annual follow-up updates. 

The test cohort consisted of 7,489 counselees. The median age was 47 and the median family size was 16. The majority (54.1\%) of counselees were recruited from population-based cancer registries. Also, 42.9\% of counselees had at least one female first-degree relative with breast cancer (a much higher proportion than in the Risk Service), 114 (1.5\%) counselees developed breast cancer within 5 years of baseline, and 1017 counselees (13.6\%) were lost to follow-up within 5 years without being diagnosed with breast cancer (Table \ref{cohorts}). To adjust for censoring, we used inverse probability of censoring weights \citep{uno2007evaluating, gerds2006consistent} (see SM \ref{appendix:ipcw} for details).

\begin{table}
\caption{Characteristics of training (Risk Service) and test (CGN) datasets.}
\centering
\begin{tabular}{llll}
  \hline
Variable & Category & Risk Service & CGN \\ 
  \hline
N (counselees) &  & 279460 & 7489 \\ 
  Age (median [IQR]) &  & 45 [39, 55] & 47 [38, 57] \\ 
  Family Size (median [IQR]) &  & 8 [7, 14] & 16 [12, 21] \\ 
  Affected 1st-degree Relatives (\%) & 0 & 242677 (86.8) & 4277 (57.1) \\ 
   & 1 & 35241 (12.6) & 2549 (34.0) \\ 
   & 2+ & 1542 (0.6) & 663 (8.9) \\ 
  Ascertainment (\%) & Population-Based & --- & 4050 (54.1) \\ 
   & Clinic-Based & --- & 2187 (29.2) \\ 
   & Self-Referral & --- & 1247 (16.7) \\ 
   & Unknown & --- & 5 (0.1) \\ 
  Censored (\%) &  & 0 (0.0) & 1017 (13.6) \\ 
  Cases (\%) &  & 13307 (4.8) & 114 (1.5) \\ 
   \hline
\end{tabular}
\label{cohorts}
\end{table}

\subsection{Training and Test Populations}

There are many differences between the Risk Service and CGN cohorts (Table \ref{cohorts}). Since CGN participants were recruited based on family history of cancer, the CGN cohort represents a higher-risk population and has more counselees with a positive family history (Table \ref{cohorts}). Due to different data collection and ascertainment procedures, the family history information available in the CGN is more detailed than in the Risk Service. To handle the considerable amount of missing age information in the Risk Service data, we did not use current or diagnosis ages of non-counselee relatives in the NN features and used only their breast and ovarian cancer affection statuses (we still used the counselee's age). 
Moreover, the Risk Service cohort is affected by selection bias because individuals who are diagnosed with breast cancer often seek genetic counseling shortly after diagnosis.  

To account for  the described  differences  between the CGN and Risk Service populations, we re-calibrated the models trained on the Risk Service to general U.S. population incidence rates adjusted for family history. 
The approaches have been previously discussed for various regression calibration problems \citep{carroll2006measurement}.
We calculated age-specific 5-year risks based on 2012-2016 incidence rates from the Surveillance, Epidemiology, and End Results (SEER) program \citep{horner2009seer}. 
We then modified the risk based on the number of affected first-degree relatives using relative risk estimates from \cite{collaborative2001familial} (the relative risks were 1.8 for 1 affected relative, 2.9 for 2 affected relatives, and 3.9 for 2 or more affected relatives). To re-calibrate each model, we used the Risk Service data to fit a linear regression with the family history-adjusted 5-year SEER risk as the outcome and the 5-year risk from the model as the predictor.  We also evaluated a re-calibrated version of BRCAPRO obtained via the SEER re-calibration approach.

\subsection{Results}
Table \ref{perf_cgn}  compares the performance of five  models (FCNN, CNN, LR, BRCAPRO, and BRCAPRO$^C$, the SEER-recalibrated version of BRCAPRO) in the CGN dataset, which was not used for training. All models underpredicted risk, with underprediction being most severe in the clinic-based subset of CGN. This may be because the CGN counselees were ascertained based on having a family history of cancer and therefore represented a higher-risk population than the sources of the data used for training and re-calibration. Overall, the NNs and BRCAPRO had comparable PR-AUCs and Brier scores, performing better than LR with respect to these metrics. The CNN and BRCAPRO also performed better than LR with respect to the AUC. In the analyses stratified by ascertainment mode, the comparisons across 1000 bootstrap replicates show evidence of accuracy improvements achieved by the CNN over the other models. Both in the  {\it  population based} test
pedigrees (63  cases)  and  in the {\it  clinic-based} test
pedigrees (39  cases),  the CNN achieved better PR-AUCs and Brier scores than LR and BRCAPRO in the  majority  of  the  bootstrap  replicates. The CNN also achieved a higher AUC than LR in the majority of the bootstrap replicates in each stratum. In the {\it  population based} pedigrees, the CNN achieved a higher AUC than BRCAPRO in 94\% of the bootstrap replicates, while in the {\it  clinic-based} pedigrees, BRCAPRO achieved a higher AUC than the CNN in 58\% of the replicates.

We performed an additional analysis where we trained the NN and LR models using only 40,000 Risk Service families instead of all 279,460 families. The models trained using the smaller sample size all performed worse (Table \ref{perf_cgn2}) than the versions trained using all Risk Service families (Table \ref{perf_cgn}). In particular, the models trained using 40,000 families had worse calibration. The FCNN had considerably lower discrimination in the overall cohort and population-based subset compared to before, indicating that large training sets are needed to develop accurate empirical models. However, the CNN trained using 40,000 families still performed reasonably well compared to BRCAPRO.


\begin{table}[!htbp]
\centering
\begingroup\footnotesize
\begin{tabular}{lllll}
  \hline
 & O/E & $\Delta$AUC & $\Delta$PR-AUC & $\Delta$sqrt(BS) \\ 
  \hline\underline{\textbf{Overall (114 cases)}} & & \\ 
 \textbf{Performance Metrics} & & \\ 
FCNN & 1.16 (0.95, 1.37) & -4.22 (-12.37, 4.76) & -6.70 (-33.06, 27.75) & -0.02 (-0.27, 0.24) \\ 
  CNN & 1.10 (0.90, 1.30) & -2.53 (-10.69, 5.92) & -4.79 (-31.35, 33.38) & 0.03 (-0.22, 0.31) \\ 
  LR & 1.07 (0.89, 1.27) & -4.56 (-12.87, 4.25) & -11.34 (-34.51, 20.35) & -0.09 (-0.36, 0.19) \\ 
  BRCAPRO & 1.34 (1.11, 1.59) & 0.00 (0.00, 0.00) & 0.00 (0.00, 0.00) & -0.03 (-0.05, -0.00) \\ 
  BRCAPRO$^C$ & 1.20 (0.99, 1.42) & AUC=0.654 & PR-AUC=0.029 & sqrt(BS)=0.130 \\ 
   \multicolumn{3}{l}{\textbf{Comparisons Across Bootstrap Replicates}} \\ 
FCNN$>$CNN & 0.090 & 0.251 & 0.334 & 0.153 \\ 
  FCNN$>$LR & 0.109 & 0.597 & 0.784 & 0.887 \\ 
  FCNN$>$BRCAPRO & 0.956 & 0.195 & 0.314 & 0.396 \\ 
  CNN$>$LR & 0.179 & 0.803 & 0.852 & 0.972 \\ 
  CNN$>$BRCAPRO & 0.923 & 0.279 & 0.386 & 0.566 \\ 
   \hline\underline{\textbf{Population-Based (63 cases)}} & & \\ 
 \textbf{Performance Metrics} & & \\ 
FCNN & 1.12 (0.87, 1.39) & 6.05 (-5.26, 16.80) & 25.25 (-14.99, 63.46) & 0.07 (-0.18, 0.28) \\ 
  CNN & 1.05 (0.81, 1.31) & 6.12 (-1.55, 14.09) & 22.35 (-12.58, 55.70) & 0.07 (-0.14, 0.26) \\ 
  LR & 1.07 (0.83, 1.34) & 4.23 (-7.14, 15.67) & 13.35 (-21.87, 53.71) & -0.04 (-0.33, 0.20) \\ 
  BRCAPRO & 1.41 (1.09, 1.76) & 0.00 (0.00, 0.00) & 0.00 (0.00, 0.00) & -0.03 (-0.05, 0.00) \\
  BRCAPRO$^C$ & 1.25 (0.97, 1.56) & AUC=0.648 & PR-AUC=0.024 & sqrt(BS)=0.128 \\ 
   \multicolumn{3}{l}{\textbf{Comparisons Across Bootstrap Replicates}} \\ 
FCNN$>$CNN & 0.269 & 0.499 & 0.618 & 0.450 \\ 
  FCNN$>$LR & 0.244 & 0.839 & 0.923 & 0.955 \\ 
  FCNN$>$BRCAPRO & 0.907 & 0.872 & 0.891 & 0.703 \\ 
  CNN$>$LR & 0.668 & 0.697 & 0.773 & 0.893 \\ 
  CNN$>$BRCAPRO & 0.864 & 0.943 & 0.891 & 0.758 \\
   \hline\underline{\textbf{Clinic-Based (39 cases)}} & & \\ 
 \textbf{Performance Metrics} & & \\ 
FCNN & 1.49 (1.08, 1.97) & -7.02 (-23.59, 14.35) & 3.08 (-44.82, 107.03) & 0.07 (-0.46, 0.67) \\ 
  CNN & 1.40 (1.01, 1.84) & -1.65 (-17.46, 18.21) & 17.62 (-37.69, 164.32) & 0.24 (-0.35, 0.90) \\ 
  LR & 1.29 (0.94, 1.71) & -5.49 (-23.12, 14.61) & -5.46 (-46.78, 62.30) & 0.05 (-0.58, 0.64) \\ 
  BRCAPRO & 1.38 (1.00, 1.84) & 0.00 (0.00, 0.00) & 0.00 (0.00, 0.00) & -0.04 (-0.07, -0.00) \\
  BRCAPRO$^C$ & 1.27 (0.92, 1.69) & AUC=0.619 & PR-AUC=0.033 & sqrt(BS)=0.146 \\ 
   \multicolumn{3}{l}{\textbf{Comparisons Across Bootstrap Replicates}} \\ 
FCNN$>$CNN & 0.016 & 0.084 & 0.024 & 0.168 \\ 
  FCNN$>$LR & 0.023 & 0.399 & 0.584 & 0.753 \\ 
  FCNN$>$BRCAPRO & 0.024 & 0.233 & 0.607 & 0.552 \\ 
  CNN$>$LR & 0.038 & 0.845 & 0.964 & 0.955 \\ 
  CNN$>$BRCAPRO & 0.046 & 0.420 & 0.803 & 0.693 \\  
   \hline
\end{tabular}
\endgroup
\caption{Performance in CGN cohort, overall and stratified by ascertainment mode. The NN and LR models were trained using a randomly selected subset of 40,000 Risk Service counselees. BRCAPRO$^C$: Re-calibrated version of BRCAPRO. $\Delta$AUC: \% relative improvement in AUC compared to BRCAPRO. $\Delta$AUC: \% relative improvement in PR-AUC compared to BRCAPRO. $\Delta$sqrt(BS): \% relative improvement in root Brier Score compared to BRCAPRO. $\rho$: correlation with BRCAPRO. In the table, the ``Comparisons Across Bootstrap Replicates" component shows pairwise comparisons between the NN models and the other models across 1000 bootstrap replicates of the test set; the row for $A>B$ shows the proportion of bootstrap replicates where model A outperformed model B with respect to each metric. } 
\label{perf_cgn}
\end{table}


\section{Discussion}\label{discussion}

The main contributions of our paper are 1) adapting FCNNs and CNNs to family history data and 2) investigating their potential for learning genetic susceptibility to cancer. To the best of our knowledge, we are the first to develop cancer risk prediction models using a dataset of more than 200,000 pedigrees. Our simulations and data application show that NNs are a promising approach for developing new models.

In simulations under the assumptions of BRCAPRO, we examined how much training data is required for NNs to achieve comparable performance to BRCAPRO. The FCNNs and CNNs trained on 200,000 or more families were highly correlated with BRCAPRO and had AUCs similar to that of BRCAPRO. With training set sizes under 200,000, the CNN performed better than the FCNN, showing that leveraging pedigree structure via convolutions can lead to more efficient training. In the setting where family history was subject to misreporting, the NNs outperformed BRCAPRO. The simulations also showed that NNs can learn feature interactions that are not pre-specified (such as rare but strongly predictive patterns involving multiple affected individuals on the same side of the family or multiple cancers in the same individual). 

In our data application, we trained NNs on over 200,000 families from the Risk Service database and validated the models on families from the CGN. In the CGN, the NNs achieved competitive performance compared to BRCAPRO in the overall cohort. They had slightly higher AUCs than BRCAPRO in population-based counselees but performed worse than BRCAPRO in clinic-based counselees with a stronger family history. These results are promising because BRCAPRO is based on domain knowledge accumulated over two decades of epidemiological studies (including \cite{miki1994strong, wooster1995identification, easton1995breast, antoniou2002comprehensive, chen2007meta}) while the NNs were trained on a single dataset. The poorer performance of the NNs in clinic-based counselees may partly be explained by the fact that the NNs used less detailed family history information than BRCAPRO. Due to missing data in the training set, we did not include age information on non-counselees in the NN inputs. This information could potentially improve the accuracy of the NNs. The performance of the NNs could also be improved by considering risk factors besides family history. Since NNs are empirical models, they can easily be extended to handle additional features by adding the features to the input vector. It is less straightforward to incorporate additional risk factors into Mendelian models because explicit assumptions need to be made about how the risk factors modify the genotype-specific risks.

Model performance can be highly dependent on how similar the test population is to the training population \citep{castaldi2011empirical, bernau2014cross}. In practice, the training and test datasets are often representative of distinct populations with different characteristics.  Some methodologies are more robust to these differences than others \citep{yu2013stability, trippa2015bayesian}. Our application is an example of  training and testing using data from different populations: the Risk Service represents a lower-risk population than the test data from the CGN, which specifically recruited participants with a family history of cancer. An advantage of training and testing in populations with different characteristics is that it allows us to evaluate how robust the model is to heterogeneity across populations. Despite the differences between the Risk Service and the CGN, the NNs trained in the Risk Service achieved comparable discriminatory accuracy to BRCAPRO, which uses parameter estimates based on higher-risk populations. Also, various methods have been developed to adjust for differences between the training and test populations \citep{janssen2008updating, sugiyama2007covariate, zhang2013domain}, which can help improve predictions.

One challenging problem we have not investigated in this paper is ascertainment, or the sampling mechanism. Pedigree-based studies of cancer risk typically use inclusion criteria that enrich for the genotypes and/or phenotypes of interest (for example, including only families with affected members). This can lead to ascertainment bias, i.e. risk estimates that are not generalizable to the population of interest. In particular, when developing pedigree-based risk prediction models, there can be differences in ascertainment between training and test datasets, and not adjusting for these differences can affect performance (especially calibration) in the test dataset. There is an extensive literature on methods for adjusting for ascertainment \citep{choi2008estimating, kraft2000bias, le1995arcad, carayol2004estimating, iversen2005population}. One approach for obtaining general population estimates from an ascertained population is to weight families by the inverse probability of being ascertained \cite{choi2008estimating}. This approach has similarities to weighting approaches that adjust for differences in covariate distributions between training and tests sets \cite{sugiyama2007covariate, zhang2013domain} and can be applied during training by using weights in the calculation of the loss function. However, the approach requires a model for the ascertainment mechanism, which is generally unknown or difficult to quantify, and is not directly applicable to existing models such as BRCAPRO. In our data application, ascertainment differed for the training and test datasets. The Risk Service counselees mostly came from mammography screening populations while the CGN counselees were ascertained based on having a family history of cancer. Moreover, there was heterogeneous ascertainment in both cohorts, since the Risk Service includes some counselees from genetic counseling clinics and the CGN used both population-based and clinic-based recruitment. We took some steps to address the ascertainment differences between the Risk Service and the CGN by re-calibrating the models trained in the Risk Service before applying them to the CGN. However, this did not perfectly calibrate the models, especially for the clinic-based subset of the CGN, highlighting the challenge of quantifying ascertainment.

While NNs allow for greater flexibility than Mendelian models and traditional regression models and do not require prior biological understanding, one disadvantage of NNs is that their black box nature makes it challenging to interpret the relationship between the predictors and risk predictions \citep{fan2020interpretability}. In contrast, traditional regression methods such as logistic regression explicitly describe monotone relationships between the predictors and risk predictions. Various post-hoc methods have been developed to determine feature importance in black box models \cite{ribeiro2016should, shrikumar2017learning}. Methods have also been proposed for developing NN models that are intrinsically interpretable \citep{dong2017towards, zhang2018interpretable, li2018deep}, but further investigation is needed in the context of family history-based cancer risk prediction.

Other disadvantages of NNs include computational burden (especially in the case of CNNs) and sample size requirements. Our simulations and data application suggest that NNs need large sample sizes ($\sim$100,000 or more) to achieve good accuracy in family history-based cancer risk prediction. In the data application, the FCNN performed particularly poorly when the training set was restricted from over 200,000 families to 40,000 families, though the CNN was still able to achieve reasonable performance. The potential benefits of using NNs are currently limited to a small number of diseases for which many pedigrees are available. In healthcare, datasets with over 100,000 pedigrees exist yet are still uncommon since collecting detailed and accurate family history is a time-consuming process. Examples besides the Risk Service include the Breakthrough Generations breast cancer study, which includes over 113,000 women \citep{swerdlow2011breakthrough}, the Swedish Family-Cancer Database, which includes over 2 million families \citep{dong2001modification}, cancer studies based on the Utah Population Database, which includes over 1.3 million probands \citep{teerlink2012comprehensive,cannon2019population}, and a cancer study based on an Icelandic genealogical database with over 600,000 individuals \citep{amundadottir2004cancer}. Though sample sizes are currently limited for most diseases, in recent years, extensive progress has been made to improve and expand family health history collection, including growing efforts in systematic data collection by research consortia \cite{john2004breast, petersen2006pancreatic, newcomb2007colon} and genetic testing companies \cite{ginsburg2019family}, the development of a wide array of electronic patient-facing family history tools \cite{Welch:2018wq}, which allow patients to gather family history information outside the clinic and therefore overcome the time constraints of traditional approaches where practitioners record family history during clinical visits, and the implementation of technology allowing for communication between family history tools and electronic health records \cite{mandel2016smart}. Also, electronic genealogical databases are rapidly expanding and there are continuing efforts to link them with clinical data to generate pedigrees \citep{stefansdottir2013use, stefansdottir2019electronically, amundadottir2004cancer, teerlink2012comprehensive}. These developments will lead to increased opportunities to refine NN models for hereditary cancer and to train NN models for other hereditary diseases.

While NNs require further development and validation before they can be considered as a viable competitor to existing family history-based models, our work indicates that they can potentially be a helpful tool for investigating and assessing familial risk.

\section*{Acknowledgements}
Work supported by the Friends of Dana-Farber Fund and NSERC PGSD35023622017. The authors thank Matthew Ploenzke for helpful suggestions.

\section*{Supplementary Material}
The supplementary material includes a notation table, an overview of standard CNNs, the proof of Theorem \ref{cnn_thm}, and additional details on the simulations and data application. The simulation code is available at \url{github.com/zoeguan/nn_cancer_risk}.


\bibliographystyle{imsart-number} 
\bibliography{refs}       

\newpage

\renewcommand\thefigure{S\arabic{figure}} 
\setcounter{figure}{0} 

\renewcommand\thetable{S\arabic{table}} 
\setcounter{table}{0} 

\appendix

\section{Methods}

\subsection{Notation}
Table \ref{table:notation} contains a summary of the notation used in the paper.

\begin{table}[H]
\footnotesize
\centering
\begin{tabular}{l|p{12cm}}
  \hline
  Variable & Description \\ 
  \hline
  $R$ & number of relatives (besides the counselee) \\ 
    $r$ & indexes family members ($r=0$ for the counselee) \\ 
    $K$ & number of cancer history features per family member \\
    $H_{ri}$ & $i$th cancer history feature for family member $r$ \\
    $A_{r1}$ & index of $r$'s mother \\
    $A_{r2}$ & indices of $r$'s father \\
    $H_r$ & $(H_{r1}, \dots, H_{rK}, A_{r1}, A_{r2})$; vector of features and parent indices for family member $r$ \\
    $H$ & family history matrix \\
    $Y_0$ & counselee's outcome \\
    & \\
    Notation for NN \\
    $Q'$ & number of relative types in reference pedigree \\
    $R_q$ & number of relatives of type $q$ in actual pedigree \\
    $R_q'$ & number of relatives of type $q$ in reference pedigree \\
    $R'$ & number of relatives in reference pedigree (besides the counselee) \\
    $H_r'$ & vector of features for family member $r$ in standardized pedigree ($K$ cancer history features plus presence/absence indicator) \\
    $H'$ & standardized version of $H$ that has the same structure as the reference pedigree \\
    $X$ & vector input to NN \\ 
    $L$ & number of hidden layers in NN \\ 
    $l$ & indexes NN layers ($l=0$ for input layer, $l=L+1$ for output layer) \\ 
    $N_l$ & number of nodes in fully-connected layer $l$\\
    $M_l$ & number of convolutional filters in layer $l$\\
    $a^l$ & output of layer $l$ \\
    $w$ & NN weight parameter \\
    $b$ & NN bias parameter \\
    $\phi$ & activation function \\
    $\sigma$ & logistic function \\
    $C$ & cost function \\
    $M$ & training sample size \\
    $\mathcal{N}(r)$ & neighborhood about member $r$, consisting of $r$ and $r$'s first-degree relatives \\
    $m_i$ & number of relatives of type $i$ ($i=1, 2, 3, 4$ for sisters, brothers, daughters, sons) in neighborhood \\
    $U$ & neighborhood size ($U=3 + \sum_{i=1}^4 m_i$) \\
   \hline
\end{tabular}
\caption{Notation table.}
 \label{table:notation}
\end{table}

\subsection{Standard CNNs}\label{appendix:cnn}

A NN is a CNN if it contains at least one convolutional layer. A convolutional layer applies the same functions repeatedly to different fixed-size regions of the layer input (for example, sets of pixels in images). These functions are called convolutional filters.

Using the same notation as in Section \ref{FCNNs} (see also Table \ref{table:notation}) unless otherwise specified, we describe a CNN that takes as input a fixed-length vector $X$ and outputs a predicted probability. Suppose the $L$ hidden layers are all convolutional layers. Let convolutional layer $l$ have $M_l$ real-valued convolutional filters $f_1^l, \dots, f_{M_l}^l$ that act on sliding windows of the input to the layer. The output of layer $l$ will be a matrix with $M_l$ columns and $N_l$ rows, where $N_l$ is the number of windows to which the $M_l$ filters are applied. $N_l$ is determined by two pre-specified quantities: the length (number of rows) of each window, which we denote by $U_l$, and the distance by which we slide (along the row axis) to move from one window to the next, which we denote by $S_l$ (typically called the stride). Therefore, $N_l = \lfloor (N_{l-1}-U_l)/S_l \rfloor$. Let $N_0$ be the length of $X$. Let $M_0=1$. 

For $i=1,\dots, M_l$, $f_i^l: \mathds{R}^{U_l * M_{l-1}} \to \mathds{R}$. Let the flattened version of the $j$th window for layer $l$ (obtained by concatenating the rows of the window) be denoted by $v_j^{l-1} \in \mathds{R}^{U_l * M_{l-1}}$. Each filter $f_i^l$ takes a weighted sum of the elements of $v_j^{l-1}$ and adds a bias term, then applies an activation function. The output from applying filter $i$ to the $j$th window is 
\[ f_i^l(v_j^{l-1}) = \phi \left(w_i^l \cdot v_j^{l-1} + b_i^l \right)\]
where $w_i^l \in \mathds{R}^{U_l * M_{l-1}}$ is the vector of weights and $b_i^l \in \mathds{R}$ is the bias for filter $i$. 

Let $f^l = (f_1^l, \dots, f_{M_l}^l): \mathds{R}^{U_l * M_{l-1}} \to \mathds{R}^{M_l}$. Then the outputs of layers $0,\dots,L$ are 
\begin{flalign*}
a^0 &= X, \\
a^l &= \left[ f^l(v_1^{l-1})^T \, | \, \dots \, | \, f^l(v_{N_l}^{l-1})^T \right]^T, \qquad (l=1,\dots,L), \\
\text{where } a^l \in \mathds{R}^{N_l \times M_l}.
\end{flalign*}

The output of the final ($l=L$) convolutional layer is flattened into a vector of length $M_L * N_L$ before being passed to the fully-connected output layer. The calculation of the prediction in the output layer and the optimization of the weight and bias parameters proceed in the same way as for FCNNs.

 When $S_l>1$, the sliding windows might not exactly cover the length of the input. To resolve this issue, CNNs are often combined with zero padding, which involves adding zeros to the borders of the layer input. We use a similar approach to apply convolutions to pedigrees (described in Section \ref{pedigreeCNNs}).

\subsection{Proof of Theorem}\label{appendix:proof}

Let $N_r^d$ be the number of relatives of $r$ of degree $d$. We first show by induction that for $d\geq0$, there exists a CNN with $d$ convolutional layers such that for any relative $r$, $a_r^d \in \mathds{R}^{(K+1)\sum_{i=0^d} N_r^i}$ ($r$'s output vector from layer $d$), is a continuous and invertible transformation of the original features $H_s'$ of $r$'s relatives $s$ of degree$\leq d$.

For $d=0$, the result holds trivially since $a_r^0 = H_r'$. Let $d>0$ and consider a CNN with $d-1$ convolutional layers that satisfies the statement for $d-1$. Any degree $d$ relative of $r$ is a degree $d-1$ relative of some first-degree relative of $r$. Therefore, by the induction assumption, there is a subset of $a_{\mathcal{N}(r)}^{d-1}$ that is a continuous and invertible transformation of the original features of relatives of $r$ of degree $\leq d$ (the subset has size $(K+1)\sum_{i=0^d} N_r^i$). Add a $d$th convolutional layer where $M_d=(K+1)\sum_{i=0^d} N_r^i$ and define the $i$th component of $f^d$ to be the application of activation function $\phi$ to the $i$th element of the subset (setting the weight for the $i$th element to 1 and the remaining weights to 0). Since $\phi$ is continuous and invertible, $a_{r}^d = f^d(a_{\mathcal{N}(r)}^{d-1})$ is a continuous, invertible transformation of the original features of relatives of $r$ of degree $\leq d$.

By the result above, there exists a CNN with $d-1$ convolutional layers such that $a_{r}^{d-1}$ is a continuous and invertible transformation of the original features of relatives of $r$ of degree $\leq d-1$. It follows that $V = a_{\mathcal{N}(0)}^{d-1} \in \mathds{R}^{U * M_{d-1}}$ (obtained by concatenating the outputs from layer $d-1$ for the counselee's neighborhood) is a continuous and invertible transformation of the original features $X$, since any degree $d$ relative is a degree $d-1$ relative of some first-degree relative. Let $tau: \mathds{R}^{U(R'+1)(K+1)} \to \mathds{R}^{M_{d-1}}$ denote the continuous and invertible transformation, so that $V = \tau(X)$.

Suppose we add another convolutional layer with $M_{d}$ filters. Then computing $a_0^{d}$ is equivalent to applying a fully-connected layer with $U M_{d}$ nodes to $V$. Since the original features are bounded and $\tau$ is continuous, $V$ is contained in a compact subset of $\mathds{R}^{U * M_{d-1}}$. The logistic function $\sigma$ is uniformly continuous on any compact subset of $\mathds{R}$, so there exists $\delta>0$ such that
\[ | X_1-X_2 |< \delta \implies | \sigma(X_1) - \sigma(X_2) |<\epsilon\]

By the universality theorem for FCNNs, there exists a value of $M_d$ and parameters for layer $d$ such that
\[ | a_0^{d} - \sigma^{-1}(g(X)) | < \delta \]
and
\[ |\sigma(a_0^{d}) - g(X)| < \epsilon \]

While the Universal Approximation Theorem for FCNN states that a given continuous function with compact support can be approximated by a single-layer FCNN with a finite number of neurons, $N$, to any degree of precision, it does not provide an upper bound for $N$. In practice, $N$ might be very large. Similarly, our theorem states that a given continuous function with compact support can be approximated by a CNN with $d$ convolutional layers, but does not provide an upper bound for the number of convolutional filters $M_d$ for the final convolutional layer $d$ (we showed above that the theorem is satisfied by a CNN with $M_d$ proportional to $N$).

\section{Simulations and Data Application}

\subsection{Misreporting Rates}\label{appendix:misreporting}

For the second simulation setting, we perturbed the family histories using the following misreporting rates for breast and ovarian cancer from \cite{ziogas2003validation} (Table \ref{table:misrep}).
\begin{table}[H]
\centering
\begin{tabular}{rrr}
  \hline
Degree & FNR & FPR \\ 
  \hline
  Breast Cancer & \\
1 & 0.05 & 0.03 \\ 
  2 & 0.18 & 0.03 \\ 
  Ovarian Cancer & \\
  1 & 0.17 & 0.01 \\ 
  2 & 0.56 & 0.02 \\ 
   \hline
\end{tabular}
\caption{Misreporting rates for breast and ovarian cancer \cite{ziogas2003validation}. FNR: False negative rate. FPR: False positive rate.}\label{table:misrep}
\end{table}

We perturbed diagnosis ages based on the rates reported in \cite{braun2017nonparametric}: 3\% of breast diagnosis ages and 4\% of ovarian cancer diagnosis ages were misreported. We assumed the difference between the true age and misreported age has mean 4 and standard deviation 3.

\subsection{Reference Pedigree Structure}\label{appendix:refped}

In the simulations, we sampled family structures from the CGN, so we chose a reference pedigree structure based on the relative counts in the CGN. We only used first- and second-degree relatives since studies have shown that excluding third- and higher-degree relatives has little effect on the performance of family history-based models \cite{biswas2013simplifying, terry201910}. In the CGN, the third quartile for the number of children of the counselee's maternal/paternal grandparents was 6, the third quartile for the number of children of the counselee's parents was 5, and the third quartile for the number of children of the counselee was 3. We used these numbers to define a symmetric family structure where each couple has an equal number of sons and daughters: the counselee's grandparents each have 3 sons and 3 daughters, the counselee's parents have 3 sons and 3 daughters, and the counselee has 2 sons and 2 daughters. We used a slightly larger family size (26) than the third quartile for family size (21) in the CGN in order to capture more of the family history. 

In the data application, we reduced the size of the reference family to 19 because the Risk Service families (median size 8, IQR 7-14) were smaller than the CGN families (median size 16, IQR 12-21). We excluded sons and daughters of the counselee from the reference structure because only 18\% of Risk Service counselees had a son or daughter in their pedigree (versus 72\% in the CGN).

\begin{figure}[!htb]
\centering
  \includegraphics[width=0.8\columnwidth]{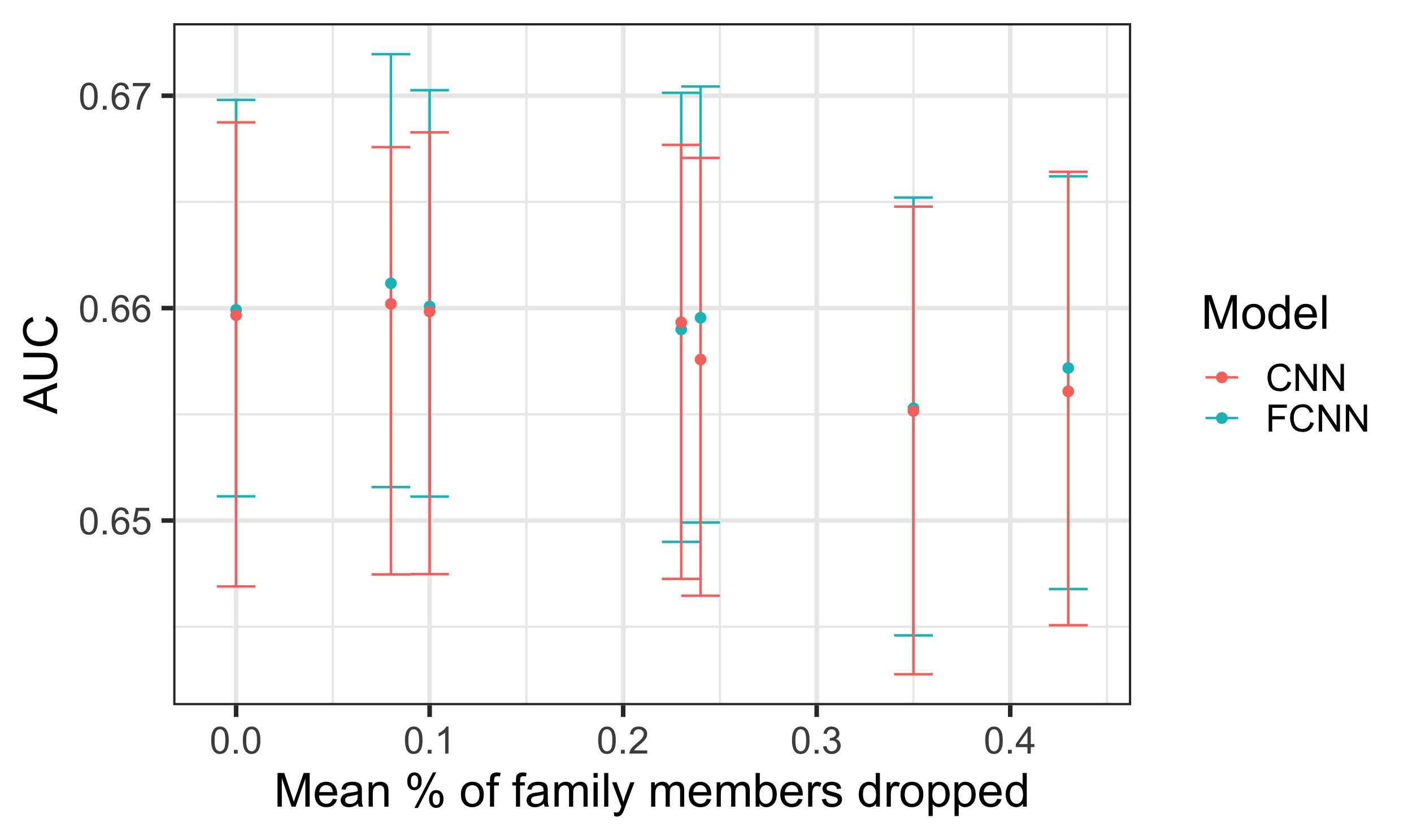}
  \caption{AUCs from sensitivity analysis for choice of reference pedigree (based on simulated data). The x-axis values (amounts of information lost from mapping the original pedigree to the reference pedigree) correspond to the reference structures defined using the counts in Table \ref{table:rel_quartiles}. The error bars represent 95\% confidence intervals based on 100 bootstrap resamples of the test set (n=87,353).}
  \label{fig:auc_refped}
\end{figure}

We performed a sensitivity analysis for the choice of reference pedigree structure using the 800,000 training families and 87,373 test families from Section \ref{sim_results} (which were simulated based on family structures from CGN data). We considered candidate reference structures based on quartiles for the number of relatives of each type (Table \ref{table:rel_quartiles}). For example, the first quartile for the number of maternal aunts in the training data is 1, so the reference structure based on the first quartile has 1 maternal aunt. In addition to the 4 reference structures based on the 4 quartiles, we also considered 3 reference structures that were constrained to be symmetric in the sense that the number of daughters is equal to the number of sons for each couple. For each of the 7 reference structures, we quantified the information lost by calculating the mean proportion of family members dropped from the original pedigree (across the training families). We also trained a FCNN and CNN for each reference structure (using the same hyperparameter values as in the main simulation analysis) and calculated the AUC. Figure \ref{fig:auc_refped} shows the performance of the NNs as a function of the amount of information lost (which ranged from 0\% for the fourth quartile to approximately 40\% for the first quartile). Overall, performance was not sensitive to the choice of reference pedigree (Figure \ref{fig:auc_refped} and Table \ref{table:refped}).

\begin{table}[!ht]
\centering
\scriptsize
\begin{tabular}{lrrrrrrr}
  \hline
relationship & Q1 & Q1s & Q2 & Q2s & Q3 & Q3s & Q4 \\ 
  \hline
Self & 1 & 1 & 1 & 1 & 1 & 1 & 1 \\ 
  Mother & 1 & 1 & 1 & 1 & 1 & 1 & 1 \\ 
  Father & 1 & 1 & 1 & 1 & 1 & 1 & 1 \\ 
  Sister & 0 & 0 & 1 & 1 & 2 & 2 & 5 \\ 
  Brother & 0 & 1 & 1 & 2 & 2 & 3 & 5 \\ 
  Daughter & 0 & 1 & 1 & 1 & 2 & 2 & 5 \\ 
  Son & 0 & 1 & 1 & 1 & 2 & 2 & 5 \\ 
  Maternal Grandmother & 1 & 1 & 1 & 1 & 1 & 1 & 1 \\ 
  Maternal Grandfather & 1 & 1 & 1 & 1 & 1 & 1 & 1 \\ 
  Paternal Grandmother & 1 & 1 & 1 & 1 & 1 & 1 & 1 \\ 
  Paternal Grandfather & 1 & 1 & 1 & 1 & 1 & 1 & 1 \\ 
  Maternal Aunt & 1 & 1 & 2 & 1 & 3 & 2 & 5 \\ 
  Maternal Uncle & 1 & 2 & 1 & 2 & 3 & 3 & 5 \\ 
  Paternal Aunt & 1 & 1 & 1 & 2 & 3 & 3 & 5 \\ 
  Paternal Uncle & 1 & 0 & 2 & 1 & 3 & 2 & 5 \\ 
   \hline
\end{tabular}
\caption{Relative counts used to define the 7 reference structures considered in our simulation-based sensitivity analysis for the choice of reference structure. Q1, Q2, Q3, and Q4 correspond to the first, second, third, and fourth quartiles of the relative counts in the simulated families, which were generated based on real family structures from the CGN (we capped the number of simulated relatives of each type at 5 to limit the computational burden). The numbers under Q1s, Q2s, and Q3s were obtained by modifying the numbers under Q1, Q2, and Q3 to enforce symmetry in the number of daughters and sons (for example, in the Q1-based reference pedigree, the counselee has no siblings, so the symmetry constraint is not satisfied, but in the Q1s-based reference pedigree, the counselee has exactly one sibling - a brother - so the number of daughters is equal to the number of sons in the counselee’s nuclear family).}\label{table:rel_quartiles}
\end{table}

\begin{table}[!htb]
\centering
\scriptsize
\begin{tabular}{llrlllll}
  \hline
Model & Q & Dropped & O/E & AUC & PR-AUC & sqrt(BS) & $\rho$ \\ 
  \hline
FCNN & Q1 & 0.43 & 0.929 (0.896, 0.964) & 0.657 (0.647, 0.666) & 0.058 (0.054, 0.064) & 0.168 (0.165, 0.171) & 0.895 (0.888, 0.902) \\ 
  FCNN & Q1s & 0.35 & 0.928 (0.895, 0.963) & 0.655 (0.645, 0.665) & 0.057 (0.053, 0.062) & 0.168 (0.165, 0.171) & 0.889 (0.884, 0.894) \\ 
  FCNN & Q2 & 0.24 & 0.941 (0.908, 0.977) & 0.66 (0.65, 0.67) & 0.059 (0.055, 0.064) & 0.168 (0.165, 0.171) & 0.899 (0.894, 0.906) \\ 
  FCNN & Q2s & 0.23 & 0.921 (0.888, 0.956) & 0.659 (0.649, 0.67) & 0.057 (0.054, 0.062) & 0.168 (0.165, 0.171) & 0.888 (0.874, 0.9) \\ 
  FCNN & Q3 & 0.08 & 0.909 (0.877, 0.944) & 0.661 (0.652, 0.672) & 0.059 (0.055, 0.065) & 0.168 (0.165, 0.171) & 0.896 (0.891, 0.903) \\ 
  FCNN & Q3s & 0.35 & 0.926 (0.893, 0.961) & 0.66 (0.651, 0.67) & 0.058 (0.054, 0.063) & 0.168 (0.165, 0.171) & 0.902 (0.89, 0.912) \\ 
  FCNN & Q4 & 0.00 & 0.904 (0.872, 0.938) & 0.66 (0.651, 0.67) & 0.057 (0.054, 0.062) & 0.168 (0.165, 0.171) & 0.879 (0.861, 0.895) \\ 
  CNN & Q1 & 0.43 & 0.997 (0.957, 1.041) & 0.656 (0.645, 0.666) & 0.058 (0.053, 0.064) & 0.168 (0.165, 0.172) & 0.892 (0.887, 0.896) \\ 
  CNN & Q1s & 0.23 & 1.009 (0.969, 1.054) & 0.655 (0.643, 0.665) & 0.057 (0.053, 0.062) & 0.168 (0.165, 0.172) & 0.893 (0.887, 0.896) \\ 
  CNN & Q2 & 0.24 & 1.021 (0.981, 1.067) & 0.658 (0.646, 0.667) & 0.059 (0.054, 0.065) & 0.168 (0.165, 0.172) & 0.907 (0.902, 0.91) \\ 
  CNN & Q2s & 0.35 & 1.004 (0.965, 1.049) & 0.659 (0.647, 0.668) & 0.058 (0.053, 0.063) & 0.168 (0.165, 0.172) & 0.907 (0.903, 0.91) \\ 
  CNN & Q3 & 0.08 & 0.991 (0.952, 1.035) & 0.66 (0.647, 0.668) & 0.06 (0.054, 0.064) & 0.168 (0.165, 0.172) & 0.918 (0.914, 0.921) \\ 
  CNN & Q3s & 0.23 & 0.994 (0.955, 1.038) & 0.66 (0.647, 0.668) & 0.059 (0.055, 0.065) & 0.168 (0.165, 0.172) & 0.917 (0.913, 0.921) \\ 
  CNN & Q4 & 0.00 & 1.01 (0.97, 1.054) & 0.66 (0.647, 0.669) & 0.06 (0.055, 0.065) & 0.168 (0.165, 0.172) & 0.911 (0.907, 0.915) \\ 
   \hline
\end{tabular}
\caption{Performance results from sensitivity analysis for choice of reference pedigree (based on simulated data). Q: quartile from Table \ref{table:rel_quartiles} used to define the reference pedigree. Dropped: mean proportion of family members dropped during pedigree mapping.}\label{table:refped}
\end{table}

\subsection{Sensitivity Analysis for Missing Data}
We performed sensitivity analyses to evaluate the impact of missing relatives and missing diagnosis ages using the simulated families (800,000 training families and 87,373 test families).  The results in Tables \ref{table:missing_train}-\ref{table:missing_test2} are very similar to the results in Table \ref{table:sim} based on the true family history with no missing data. For the missing data sensitivity analyses, we first considered scenarios where there was only missing data in the training set (Tables \ref{table:missing_train} and \ref{table:missing_train2}) as well as scenarios where there was only missing data in the test set (Tables \ref{table:missing_test} and \ref{table:missing_test2}). We varied the proportion of missing relatives (Tables \ref{table:missing_train} and \ref{table:missing_test}) or missing diagnosis ages (Tables \ref{table:missing_train2} and \ref{table:missing_test2}) from 0.05 to 0.3. For example, when the parameter was 0.05, we randomly removed 5\% of the relatives in the simulated pedigrees and we randomly set 5\% of the diagnosis ages of affected relatives as missing. We considered removing relatives without considering their affection status as well as removing only unaffected relatives. Test set performance was similar across the scenarios (Tables \ref{table:missing_train}-\ref{table:missing_test2}). The NNs tended to perform slightly worse in scenarios where 30\% of relatives were missing from the test set (Table \ref{table:missing_test}), as expected, but the differences were small. For example, the AUC of the FCNN decreased from 0.658 to 0.646 when the proportion of missing relatives in the test set increased from 5\% to 30\%. 

\begin{table}[ht]
\centering
\scriptsize
\begin{tabular}{lrlllll}
  \hline
Model & Missing & O/E & AUC & PR-AUC & sqrt(BS) & $\rho$ \\ 
  \hline
\multicolumn{4}{l}{\textbf{Randomly Missing Relatives}} & \\
FCNN & 0.05 & 0.918 (0.885, 0.952) & 0.66 (0.651, 0.67) & 0.058 (0.054, 0.063) & 0.168 (0.165, 0.171) & 0.886 (0.872, 0.898) \\ 
  FCNN & 0.10 & 0.933 (0.9, 0.969) & 0.661 (0.652, 0.671) & 0.058 (0.054, 0.062) & 0.168 (0.165, 0.171) & 0.905 (0.899, 0.911) \\ 
  FCNN & 0.30 & 0.936 (0.903, 0.971) & 0.659 (0.65, 0.67) & 0.058 (0.055, 0.064) & 0.168 (0.165, 0.171) & 0.892 (0.884, 0.901) \\ 
  CNN & 0.05 & 0.989 (0.951, 1.033) & 0.66 (0.647, 0.668) & 0.059 (0.054, 0.064) & 0.168 (0.165, 0.172) & 0.916 (0.913, 0.92) \\ 
  CNN & 0.10 & 1 (0.961, 1.045) & 0.658 (0.646, 0.667) & 0.059 (0.054, 0.064) & 0.168 (0.165, 0.172) & 0.912 (0.909, 0.916) \\ 
  CNN & 0.30 & 0.971 (0.932, 1.014) & 0.659 (0.648, 0.668) & 0.059 (0.054, 0.064) & 0.168 (0.165, 0.172) & 0.911 (0.908, 0.915) \\ 
\hline
\multicolumn{4}{l}{\textbf{Randomly Missing Unaffected Relatives}} & \\
  FCNN & 0.05 & 0.935 (0.902, 0.97) & 0.66 (0.651, 0.67) & 0.058 (0.054, 0.062) & 0.168 (0.165, 0.171) & 0.886 (0.87, 0.9) \\ 
  FCNN & 0.10 & 0.927 (0.894, 0.962) & 0.66 (0.651, 0.67) & 0.058 (0.054, 0.063) & 0.168 (0.165, 0.171) & 0.892 (0.876, 0.906) \\ 
  FCNN & 0.30 & 0.942 (0.908, 0.977) & 0.661 (0.652, 0.671) & 0.057 (0.054, 0.061) & 0.168 (0.165, 0.171) & 0.821 (0.791, 0.847) \\ 
  CNN & 0.05 & 0.994 (0.955, 1.038) & 0.66 (0.648, 0.668) & 0.059 (0.054, 0.064) & 0.168 (0.165, 0.172) & 0.917 (0.913, 0.921) \\ 
  CNN & 0.10 & 1.015 (0.975, 1.06) & 0.66 (0.647, 0.668) & 0.059 (0.054, 0.064) & 0.168 (0.165, 0.172) & 0.914 (0.91, 0.918) \\ 
  CNN & 0.30 & 0.99 (0.951, 1.034) & 0.659 (0.647, 0.667) & 0.059 (0.054, 0.064) & 0.168 (0.165, 0.172) & 0.912 (0.909, 0.916) \\ 
   \hline
\end{tabular}
\caption{Test set performance for simulation scenarios where relatives were randomly removed from the training set. Missing: proportion of relatives randomly removed. O/E: observed to expected number of cases. AUC: area under the receiver operating characteristic curve. PR-AUC: area under the precision-recall curve. sqrt(BS): square root of Brier Score. $\rho$: correlation with true model (BRCAPRO).}\label{table:missing_train}
\end{table}

\begin{table}[ht]
\centering
\scriptsize
\begin{tabular}{rlrlllll}
  \hline
 Model & Missing & OE & AUC & PR-AUC & sqrt(BS) & $\rho$\\ 
  \hline
FCNN & 0.05 & 0.928 (0.895, 0.963) & 0.66 (0.651, 0.67) & 0.058 (0.054, 0.063) & 0.168 (0.165, 0.171) & 0.902 (0.89, 0.911) \\ 
  FCNN & 0.10 & 0.931 (0.898, 0.966) & 0.66 (0.651, 0.67) & 0.058 (0.054, 0.063) & 0.168 (0.165, 0.171) & 0.901 (0.894, 0.909) \\ 
  FCNN & 0.30 & 0.936 (0.903, 0.972) & 0.66 (0.651, 0.67) & 0.058 (0.054, 0.062) & 0.168 (0.165, 0.171) & 0.9 (0.894, 0.906) \\ 
  CNN & 0.05 & 0.994 (0.955, 1.038) & 0.66 (0.647, 0.668) & 0.06 (0.054, 0.064) & 0.168 (0.165, 0.172) & 0.916 (0.913, 0.92) \\ 
  CNN & 0.10 & 0.996 (0.957, 1.04) & 0.66 (0.647, 0.668) & 0.059 (0.054, 0.064) & 0.168 (0.165, 0.172) & 0.916 (0.912, 0.92) \\ 
  CNN & 0.30 & 0.995 (0.956, 1.039) & 0.66 (0.647, 0.668) & 0.059 (0.054, 0.064) & 0.168 (0.165, 0.172) & 0.915 (0.912, 0.919) \\ 
   \hline
\end{tabular}
\caption{Test set performance for simulation scenarios with missing diagnosis ages in the training set. Missing: proportion of missing diagnosis ages among affected relatives. O/E: observed to expected number of cases. AUC: area under the receiver operating characteristic curve. PR-AUC: area under the precision-recall curve. sqrt(BS): square root of Brier Score. $\rho$: correlation with true model (BRCAPRO).}\label{table:missing_train2}
\end{table}

\begin{table}[ht]
\centering
\scriptsize
\begin{tabular}{lrlllll}
  \hline
Model & Missing & O/E & AUC & PR-AUC & sqrt(BS) & $\rho$ \\ 
  \hline
\multicolumn{4}{l}{\textbf{Randomly Missing Relatives}} & \\
FCNN & 0.05 & 0.925 (0.89, 0.965) & 0.658 (0.648, 0.67) & 0.058 (0.053, 0.064) & 0.168 (0.165, 0.172) & 0.889 (0.875, 0.899) \\ 
  FCNN & 0.10 & 0.924 (0.892, 0.961) & 0.656 (0.644, 0.665) & 0.058 (0.054, 0.064) & 0.168 (0.165, 0.171) & 0.88 (0.872, 0.887) \\ 
  FCNN & 0.30 & 0.917 (0.883, 0.944) & 0.646 (0.638, 0.655) & 0.055 (0.052, 0.06) & 0.168 (0.165, 0.17) & 0.832 (0.82, 0.84) \\ 
  CNN & 0.05 & 1.006 (0.975, 1.033) & 0.659 (0.649, 0.667) & 0.059 (0.055, 0.065) & 0.168 (0.165, 0.17) & 0.908 (0.904, 0.912) \\ 
  CNN & 0.10 & 1.018 (0.985, 1.057) & 0.656 (0.645, 0.666) & 0.059 (0.055, 0.067) & 0.168 (0.165, 0.171) & 0.899 (0.894, 0.903) \\ 
  CNN & 0.30 & 1.07 (1.036, 1.11) & 0.648 (0.637, 0.658) & 0.057 (0.053, 0.062) & 0.168 (0.166, 0.171) & 0.863 (0.859, 0.869) \\
\hline
\multicolumn{4}{l}{\textbf{Randomly Missing Unaffected Relatives}} & \\
  FCNN & 0.05 & 0.918 (0.883, 0.955) & 0.658 (0.647, 0.666) & 0.057 (0.053, 0.062) & 0.168 (0.165, 0.171) & 0.897 (0.884, 0.906) \\ 
  FCNN & 0.10 & 0.91 (0.886, 0.938) & 0.658 (0.647, 0.672) & 0.058 (0.054, 0.063) & 0.168 (0.166, 0.171) & 0.892 (0.879, 0.9) \\ 
  FCNN & 0.30 & 0.878 (0.844, 0.914) & 0.656 (0.647, 0.667) & 0.058 (0.054, 0.064) & 0.168 (0.165, 0.171) & 0.884 (0.879, 0.888) \\ 
  CNN & 0.05 & 0.997 (0.957, 1.026) & 0.66 (0.65, 0.668) & 0.059 (0.055, 0.066) & 0.168 (0.165, 0.171) & 0.916 (0.913, 0.92) \\ 
  CNN & 0.10 & 0.999 (0.966, 1.026) & 0.661 (0.652, 0.671) & 0.06 (0.056, 0.066) & 0.168 (0.166, 0.17) & 0.916 (0.912, 0.919) \\ 
  CNN & 0.30 & 1.011 (0.975, 1.043) & 0.66 (0.65, 0.669) & 0.06 (0.055, 0.066) & 0.168 (0.165, 0.171) & 0.914 (0.909, 0.918) \\
   \hline
\end{tabular}
\caption{Test set performance for simulation scenarios where relatives were randomly removed from the test set. Missing: proportion of missing diagnosis ages among affected relatives. O/E: observed to expected number of cases. AUC: area under the receiver operating characteristic curve. PR-AUC: area under the precision-recall curve. sqrt(BS): square root of Brier Score. $\rho$: correlation with true model (BRCAPRO).}\label{table:missing_test}
\end{table}

\begin{table}[ht]
\centering
\scriptsize
\begin{tabular}{rlrlllll}
  \hline
Model & Missing & OE & AUC & PR-AUC & sqrt(BS) & $\rho$ \\ 
  \hline
FCNN & 0.05 & 0.926 (0.892, 0.961) & 0.66 (0.652, 0.67) & 0.058 (0.054, 0.063) & 0.168 (0.165, 0.171) & 0.902 (0.889, 0.911) \\ 
  FCNN & 0.10 & 0.925 (0.897, 0.961) & 0.66 (0.653, 0.67) & 0.058 (0.055, 0.063) & 0.168 (0.166, 0.171) & 0.901 (0.889, 0.91) \\ 
  FCNN & 0.30 & 0.923 (0.899, 0.95) & 0.659 (0.648, 0.668) & 0.058 (0.053, 0.062) & 0.168 (0.166, 0.171) & 0.899 (0.885, 0.908) \\ 
  CNN & 0.05 & 0.993 (0.954, 1.037) & 0.66 (0.647, 0.668) & 0.059 (0.054, 0.064) & 0.168 (0.165, 0.172) & 0.916 (0.913, 0.92) \\ 
  CNN & 0.10 & 0.992 (0.96, 1.03) & 0.66 (0.651, 0.671) & 0.059 (0.055, 0.066) & 0.168 (0.165, 0.171) & 0.916 (0.912, 0.919) \\ 
  CNN & 0.30 & 0.99 (0.962, 1.035) & 0.658 (0.65, 0.668) & 0.059 (0.055, 0.064) & 0.168 (0.166, 0.172) & 0.914 (0.911, 0.918) \\ 
   \hline
\end{tabular}
\caption{Test set performance for simulation scenarios with missing diagnosis ages in the test set. Missing: proportion of missing diagnosis ages among affected relatives. O/E: observed to expected number of cases. AUC: area under the receiver operating characteristic curve. PR-AUC: area under the precision-recall curve. sqrt(BS): square root of Brier Score. $\rho$: correlation with true model (BRCAPRO).}\label{table:missing_test2}
\end{table}

\subsection{Choice of Loss Function}\label{appendix:loss}

In addition to the equivalence between mean squared error loss and the Brier score, one justification for using mean squared error loss in training NNs for predicting binary outcomes is that, like cross-entropy loss, mean squared error loss has a valid probabilistic interpretation \cite{hampshire1991equivalence, janocha2017loss}. Also, a recent paper \cite{hui2020evaluation} showed empirically that mean square error loss performs comparably or better than cross-entropy loss in many classification problems. However, mean square error loss can lead to slower convergence \cite{janocha2017loss}. To check whether our results were sensitive to the choice of the loss function, we repeated the main simulation analysis using cross-entropy loss instead of mean squared error loss. We found that there were minimal differences based on standard performance metrics (Table \ref{table:loss}).

\begin{table}[!ht]
\centering
\begin{tabular}{rlllll}
  \hline
 & OE & AUC & PR-AUC & sqrt(BS) & $\rho$ \\ 
  \hline
FCNN MSE & 0.93 (0.89, 0.961) & 0.66 (0.651, 0.67) & 0.058 (0.054, 0.063) & 0.168 (0.165, 0.171) & 0.9 (0.89, 0.91) \\ 
  FCNN CE & 0.9 (0.87, 0.938) & 0.657 (0.647, 0.667) & 0.059 (0.055, 0.064) & 0.168 (0.165, 0.171) & 0.92 (0.91, 0.92) \\ 
  CNN MSE & 0.99 (0.95, 1.038) & 0.66 (0.647, 0.668) & 0.059 (0.055, 0.065) & 0.168 (0.165, 0.172) & 0.92 (0.91, 0.92) \\ 
  CNN CE & 0.94 (0.91, 0.985) & 0.659 (0.647, 0.667) & 0.059 (0.054, 0.064) & 0.168 (0.165, 0.172) & 0.92 (0.91, 0.92) \\ 
   \hline
\end{tabular}
\caption{Performance of NNs trained using mean squared error (MSE) versus cross-entropy (CE) loss in simulated families (training set of 800,000). FCNN MSE: FCNN trained using MSE loss. FCNN CE: FCNN trained using CE loss. CNN MSE: CNN trained using MSE loss. O/E: observed to expected number of cases. AUC: area under the receiver operating characteristic curve. PR-AUC: area under the precision-recall curve. sqrt(BS): square root of Brier Score. $\rho$: correlation with true model (BRCAPRO).}\label{table:loss}
\end{table}

\subsection{Precision-Recall Curves} 
\begin{figure}[!htb]
\centering
  \includegraphics[width=0.8\columnwidth]{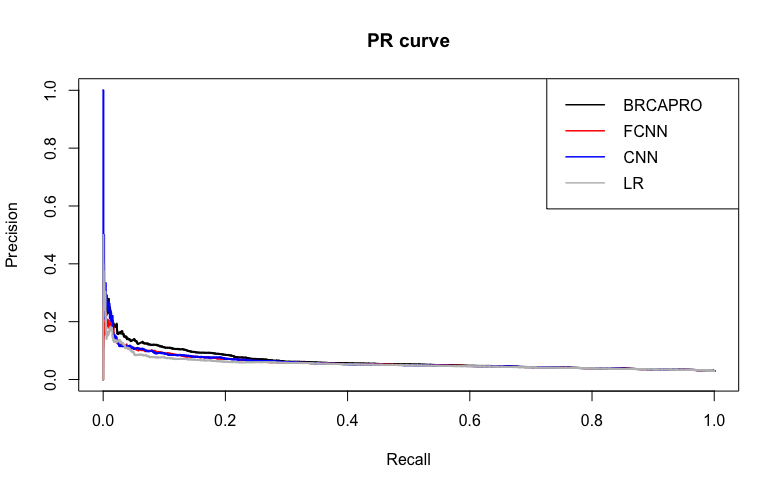}
  \caption{Precision-recall curves for the main simulation scenario.}
  \label{fig:pr_curve}
\end{figure}

Figure \ref{fig:pr_curve} shows the precision-recall curves of the models from the main simulation scenario. The precision-recall curve is highly sensitive to disease prevalence (unlike ROC-AUC). The proportion of cases in our dataset is very small therefore all of the models have low PR-AUCs (the baseline PR-AUC, or PR-AUC of a model that does no better than random guessing, is the proportion of cases, 0.029).

\subsection{Training Time} 
\begin{figure}[!htb]
\centering
  \includegraphics[width=0.8\columnwidth]{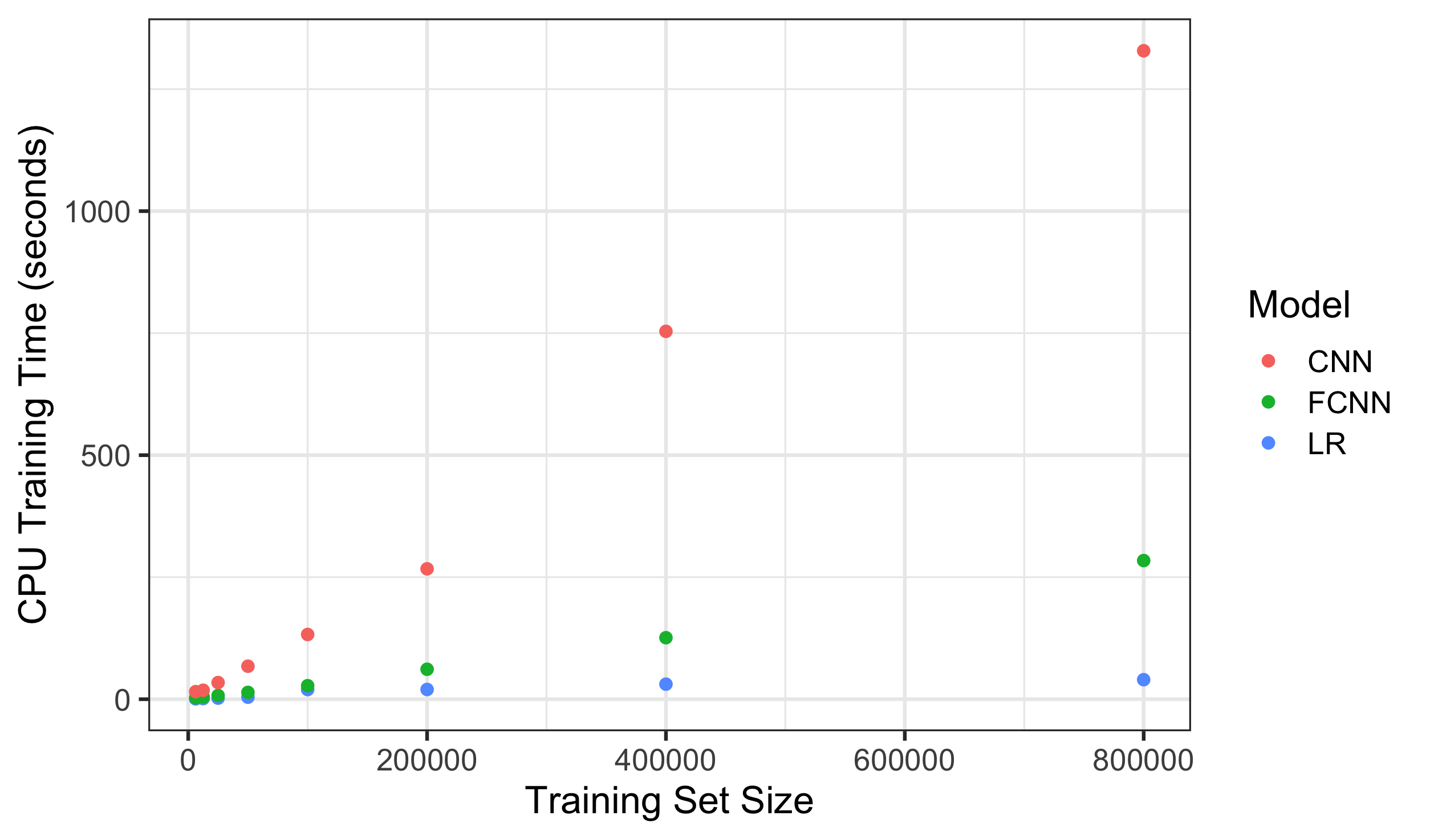}
  \caption{Training time as a function of training sample size. All models were trained using a single CPU core. The NNs were trained using the same hyperparameters as in the main simulation analysis. The FCNNs were trained for 30 epochs and the CNNs were trained for 15 epochs.}
  \label{fig:training_time}
\end{figure}

Figure \ref{fig:training_time} shows the computational intensity of training FCNN, CNN, and LR models on the simulated families.

\subsection{Missing Age Information in the Risk Service}\label{appendix:missingages}

Ages and diagnoses ages were available for all but 185 counselees, who were excluded from the training set. If an affected first- or second-degree relative was missing age or age at diagnosis, then we set their baseline cancer status as affected. One justification for this imputation approach is that counselees may have more accurate memory of recent diagnoses (i.e. within the last 5 years) among relatives than less recent diagnoses. Therefore, the missing diagnoses ages are more likely to correspond to less recent diagnoses. 

We considered including current age and age at diagnosis in the models trained on the Risk Service (and adding a missing indicator for each age variable), but in cross-validation on the Risk Service, the models without age information performed better than the models with age information, so we excluded the age information (other than the counselee's current age).

\subsection{Inverse Probability of Censoring Weights}\label{appendix:ipcw}

To account for censoring, individuals with observed outcomes are used to calculate the performance measures and are weighted by their inverse probability of not being censored by the minimum of 1) the length of the risk prediction period (5 years) and 2) the time to breast cancer diagnosis. Individuals who are censored are not directly used to calculate the performance measures, but are used to estimate the censoring distribution. We assumed independent censoring and estimated the censoring distribution using the Kaplan-Meier estimator.

\subsection{Data Application - Smaller Training Set}

To evaluate the performance of the NNs in a real data setting with a smaller training sample size than $\sim$ 200,000, we reran the data application after restricting the Risk Service training dataset to 40,000 randomly selected families. The CNN and LR models trained using 40,000 families (Table \ref{perf_cgn2}) performed slightly worse than those trained using all 279,460 families (Table \ref{perf_cgn}), while the FCNN performed considerably worse.

\begin{table}[!htbp]
\centering
\begingroup\footnotesize
\begin{tabular}{lllll}
  \hline
 & O/E & $\Delta$AUC & $\Delta$PR-AUC & $\Delta$sqrt(BS) \\ 
  \hline\underline{\textbf{Overall (114 cases)}} & & \\ 
 \textbf{Performance Metrics} & & \\ 
FCNN & 1.26 (1.04, 1.49) & -10.49 (-19.81, -0.43) & -15.80 (-40.93, 18.53) & -0.09 (-0.34, 0.16) \\ 
  CNN & 1.20 (0.98, 1.42) & -5.67 (-14.52, 3.36) & -6.31 (-33.29, 29.43) & -0.03 (-0.28, 0.21) \\ 
  LR & 1.24 (1.02, 1.47) & -6.87 (-16.38, 2.90) & -10.30 (-36.08, 25.30) & -0.09 (-0.38, 0.19) \\ 
  BRCAPRO & 1.34 (1.11, 1.59) & 0.00 (0.00, 0.00) & 0.00 (0.00, 0.00) & -0.03 (-0.05, -0.00) \\ 
  BRCAPRO$^C$ & 1.20 (0.99, 1.42) & AUC=0.654 & PR-AUC=0.029 & sqrt(BS)=0.130 \\ 
   \multicolumn{4}{l}{\textbf{Comparisons Across Bootstrap Replicates}} \\ 
FCNN$>$CNN & 0.018 & 0.006 & 0.071 & 0.107 \\ 
  FCNN$>$LR & 0.013 & 0.033 & 0.164 & 0.557 \\ 
  FCNN$>$BRCAPRO & 0.018 & 0.022 & 0.160 & 0.235 \\ 
  CNN$>$LR & 0.977 & 0.749 & 0.820 & 0.903 \\ 
  CNN$>$BRCAPRO & 0.627 & 0.113 & 0.323 & 0.382 \\ 
   \hline\underline{\textbf{Population-Based (63 cases)}} & & \\ 
 \textbf{Performance Metrics} & & \\ 
FCNN & 1.22 (0.94, 1.52) & -3.87 (-16.07, 9.19) & 5.88 (-29.68, 49.42) & -0.06 (-0.25, 0.14) \\ 
  CNN & 1.14 (0.88, 1.42) & 3.45 (-5.99, 13.87) & 28.93 (-13.81, 73.82) & 0.04 (-0.16, 0.24) \\ 
  LR & 1.19 (0.92, 1.48) & -0.18 (-10.92, 12.07) & 14.65 (-20.92, 49.17) & -0.08 (-0.30, 0.13) \\ 
  BRCAPRO & 1.41 (1.09, 1.76) & 0.00 (0.00, 0.00) & 0.00 (0.00, 0.00) & -0.03 (-0.05, 0.00) \\
  BRCAPRO$^C$ & 1.25 (0.97, 1.56) & AUC=0.648 & PR-AUC=0.024 & sqrt(BS)=0.128 \\ 
   \multicolumn{4}{l}{\textbf{Comparisons Across Bootstrap Replicates}} \\ 
FCNN$>$CNN & 0.105 & 0.006 & 0.020 & 0.050 \\ 
  FCNN$>$LR & 0.081 & 0.069 & 0.184 & 0.634 \\ 
  FCNN$>$BRCAPRO & 0.948 & 0.251 & 0.557 & 0.236 \\ 
  CNN$>$LR & 0.884 & 0.923 & 0.966 & 0.991 \\ 
  CNN$>$BRCAPRO & 0.916 & 0.768 & 0.897 & 0.644 \\   
   \hline\underline{\textbf{Clinic-Based (39 cases)}} & & \\ 
 \textbf{Performance Metrics} & & \\ 
FCNN & 1.63 (1.18, 2.15) & -2.35 (-20.53, 21.67) & 2.22 (-42.96, 104.98) & 0.07 (-0.45, 0.68) \\ 
  CNN & 1.56 (1.14, 2.07) & -2.56 (-19.51, 18.26) & 5.70 (-41.68, 90.05) & 0.10 (-0.44, 0.70) \\ 
  LR & 1.65 (1.19, 2.18) & -1.07 (-18.59, 21.61) & 9.30 (-39.40, 99.19) & 0.14 (-0.47, 0.81) \\
  BRCAPRO & 1.38 (1.00, 1.84) & 0.00 (0.00, 0.00) & 0.00 (0.00, 0.00) & -0.04 (-0.07, -0.00) \\
  BRCAPRO$^C$ & 1.27 (0.92, 1.69) & AUC=0.619 & PR-AUC=0.033 & sqrt(BS)=0.146 \\ 
   \multicolumn{4}{l}{\textbf{Comparisons Across Bootstrap Replicates}} \\ 
FCNN$>$CNN & 0.002 & 0.509 & 0.376 & 0.443 \\ 
  FCNN$>$LR & 0.889 & 0.341 & 0.308 & 0.344 \\ 
  FCNN$>$BRCAPRO & 0.014 & 0.397 & 0.606 & 0.563 \\ 
  CNN$>$LR & 0.998 & 0.243 & 0.343 & 0.345 \\ 
  CNN$>$BRCAPRO & 0.018 & 0.389 & 0.663 & 0.596 \\   
   \hline
\end{tabular}
\endgroup
\caption{Performance in CGN cohort, overall and stratified by ascertainment mode, for NN and LR models trained using a randomly selected subset of 40,000 Risk Service counselees. The performance metrics for BRCAPRO are the same as in Table 3 but are included for comparison. BRCAPRO$^C$: Re-calibrated version of BRCAPRO. $\Delta$AUC: \% relative improvement in AUC compared to BRCAPRO. $\Delta$AUC: \% relative improvement in PR-AUC compared to BRCAPRO. $\Delta$sqrt(BS): \% relative improvement in root Brier Score compared to BRCAPRO. In the table, the ``Comparisons Across Bootstrap Replicates" component shows pairwise comparisons between the NN models and the other models across 1000 bootstrap replicates of the test set; the row for $A>B$ shows the proportion of bootstrap replicates where model A outperformed model B with respect to each metric.} 
\label{perf_cgn2}
\end{table}

\end{document}